\documentclass[11pt]{article}

\usepackage[preprint]{acl}
\usepackage{times}
\usepackage{latexsym}
\usepackage{natbib}

\usepackage[T1]{fontenc}
\usepackage[utf8]{inputenc}

\usepackage{microtype}

\usepackage{inconsolata}

\usepackage{graphicx}
\usepackage{booktabs}
\usepackage{multirow}
\usepackage{xcolor}
\usepackage{booktabs}
\usepackage{soul}      % 导言区
\usepackage{amsmath}
\usepackage{subcaption} % very important
\sethlcolor{yellow}    % 可改成别的颜色
\usepackage[table]{xcolor}
\usepackage{makecell}
\usepackage{multirow,array}
\usepackage{multirow,booktabs}
\usepackage{adjustbox}
\usepackage{enumitem}
\usepackage{fancyvrb}
\usepackage{fvextra} % 推荐：自带 fancyvrb，并支持 breaklines
\usepackage{float}
\usepackage{amsfonts}  
\usepackage{tabularx}
\usepackage{hyperref}

\title{CtrlCoT: Dual-Granularity Chain-of-Thought Compression for Controllable Reasoning}

\author{
 \textbf{Zhenxuan Fan\textsuperscript{1}},
 \textbf{Jie Cao\textsuperscript{1}},
 \textbf{Yang Dai\textsuperscript{1}},
 \textbf{Zheqi Lv\textsuperscript{1}},
 \textbf{Wenqiao Zhang\textsuperscript{1}\thanks{Corresponding author}},
\\
 \textbf{Zhongle Xie\textsuperscript{1}},
 \textbf{Peng LU\textsuperscript{1}},
 \textbf{Beng Chin Ooi \textsuperscript{1}},
\\
 \textsuperscript{1}Zhejiang University
\\
}

\begin{document}
\maketitle

\begin{abstract}

Chain-of-thought (CoT) prompting improves LLM reasoning but incurs high latency and memory cost due to verbose traces, motivating CoT compression with preserved correctness. Existing methods either shorten CoTs at the semantic level, which is often conservative, or prune tokens aggressively, which can miss task-critical cues and degrade accuracy. Moreover, combining the two is non-trivial due to sequential dependency, task-agnostic pruning, and distribution mismatch.
We propose \textbf{CtrlCoT}, a dual-granularity CoT compression framework that harmonizes semantic abstraction and token-level pruning through three components: Hierarchical Reasoning Abstraction produces CoTs at multiple semantic granularities; Logic-Preserving Distillation trains a logic-aware pruner to retain indispensable reasoning cues (e.g., numbers and operators) across pruning ratios; and Distribution-Alignment Generation aligns compressed traces with fluent inference-time reasoning styles to avoid fragmentation.
On MATH-500 with Qwen2.5-7B-Instruct, CtrlCoT uses 30.7\% fewer tokens while achieving 7.6 percentage points higher than the strongest baseline, demonstrating more efficient and reliable reasoning. Our code will be publicly available at \url{https://github.com/fanzhenxuan/Ctrl-CoT}.
\end{abstract}

% lvzheqi：图的坐标轴可以保持不变，但纵向可以小一点（比如缩1/4）。你应该是matplotlib画的吧，那样的话有个plt.figure()那里把纵向的那个数值调低一点就行。如果你怕调低之后有的坐标显不出来，可以等比例调，调完反正你这都导入也不会改变他的纵横比，
% zx:宽度没变，高度已从15调到12

\section{Introduction}
% Chain-of-thought (CoT) prompting~\citep{NEURIPS2022_9d560961} has driven a substantial leap in large language models (LLMs)~\citep{openai2024openaio1card, deepseekai2025deepseekr1incentivizingreasoningcapability, comanici2025gemini25pushingfrontier}, enabling remarkable performance on complex reasoning tasks such as arithmetic, logical deduction, and code generation~\citep{DBLP:conf/iclr/LuoSX0LTGLCT025,wang2025logictree,ma2025rethinking,yu2025thinkrec,lv2025multimodal}. 

% Large language models (LLMs) and multimodal large language models (MLLMs) have progressed rapidly, enabling broad real-world applications across diverse domains~\citep{lin2025healthgpt, tang-etal-2025-financereasoning, yuan2025videorefer,xie2025heartcare}. 
With the rapid advancement of deep learning technologies~\citep{brown2020language,zhang2022boostmis}, 
large language models (LLMs) and multimodal large language models (MLLMs) have progressed rapidly, enabling broad real-world applications across diverse domains~\citep{lin2025healthgpt, yuan2025videorefer}.
In parallel, Chain-of-thought (CoT) prompting~\citep{NEURIPS2022_9d560961} has emerged as an effective paradigm for enhancing LLM reasoning, leading to substantial improvements on complex tasks across different domains~\citep{DBLP:conf/iclr/LuoSX0LTGLCT025, wang2025logictree, yu2025thinkrec}.
% However, this performance gain comes at a cost: generating verbose intermediate traces significantly increases autoregressive decoding time and key-value cache usage, leading to high latency and serving costs~\citep{chen2025do,fan2025missing}.
However, this performance gain comes at a cost: verbose intermediate traces increase decoding latency and serving overhead~\citep{chen2025do,fan2025missing}.
Consequently, compressing CoT while maintaining reasoning fidelity has become a critical research 
direction~\citep{sui2025stopoverthinkingsurveyefficient}.
% zhu2025towards}
\begin{figure}[t]       % [t]=顶部 [h]=当前位置 [b]=底部 [!ht]=尽量当前位置
  \centering
  \includegraphics[width=\linewidth]{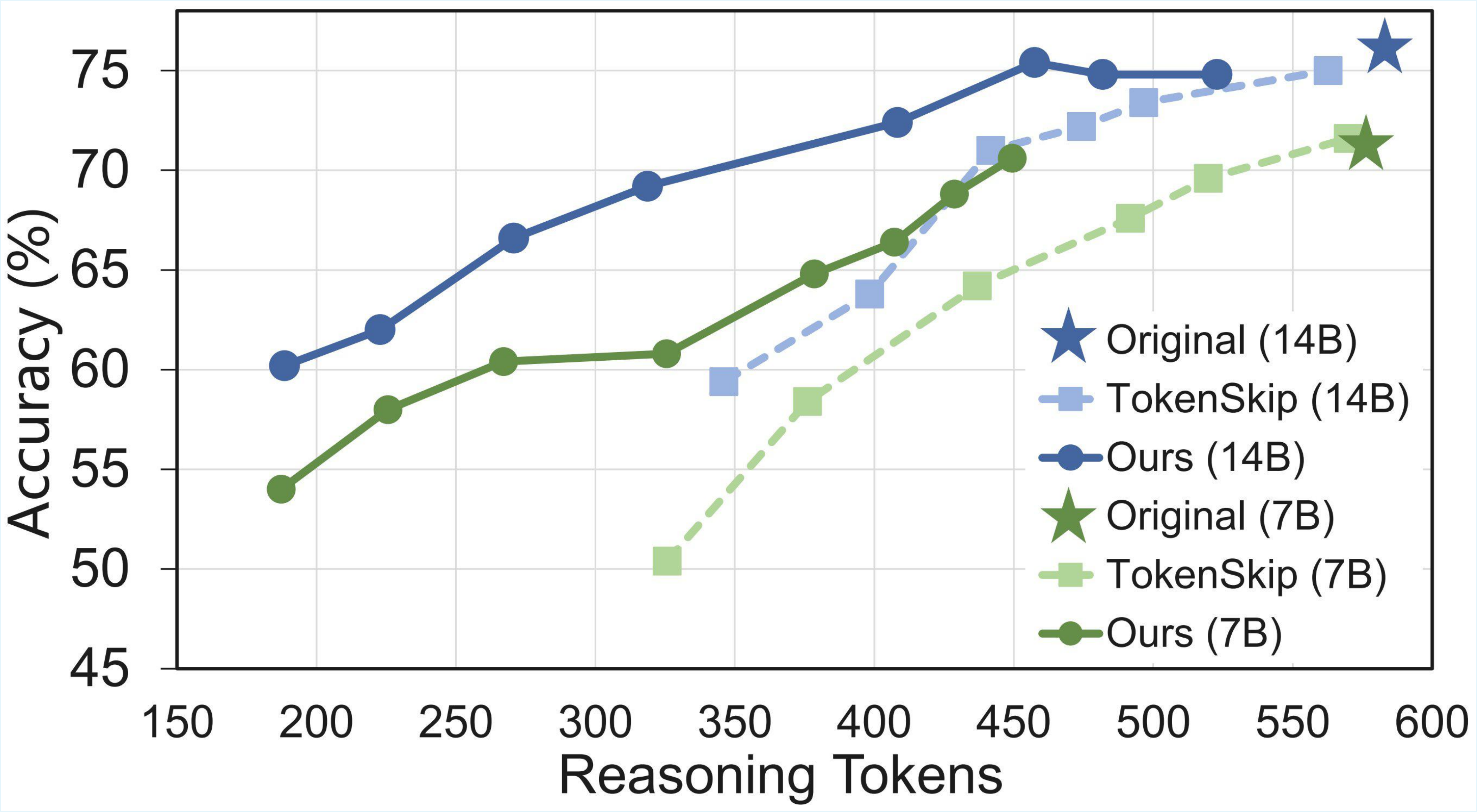} % 也可以写成 width=0.8\linewidth
\vspace{-0.75cm}
\caption{Accuracy versus CoT length on MATH-500 under different token budgets for Qwen2.5-7B/14B-Instruct, where our method consistently achieves higher accuracy at comparable CoT lengths.}
  \label{fig:example}
\vspace{-0.5cm}
\end{figure}

Existing compression methods generally fall into two categories: semantic-level shortening and token-level skipping~\citep{han2025token,ma2025reasoning,muennighoff2025s1simpletesttimescaling}. However, each of them inherently suffers from limitations, making it difficult to balance compression ratio and accuracy:
(i) \textit{Semantic-level approaches} achieve simplification by rewriting reasoning trajectories, but their compression potential is constrained by the requirement of semantic integrity. 
For instance, LC-Prompt~\citep{xia2025tokenskip} 
% preserves accuracy on \textsc{MATH-500} (71.2\%$\rightarrow$71.0\%)
nearly maintains the original accuracy on the MATH-500 dataset (71.2\% → 71.0\%) but only achieves a compression ratio of 11\%, indicating substantial room for optimization under its conservative strategy;
(ii) \textit{Token-level approaches} can achieve higher compression ratios, they significantly degrade model performance due to the lack of semantic understanding.
Under the same experimental setup, TokenSkip~\citep{xia2025tokenskip} reduces the number of tokens by 43\% but causes an accuracy drop of over 20 percentage points, a performance degradation that limits its practical applications.

While semantic- and token-level methods are complementary, combining them is not plug-and-play: naïve integration raises technical challenges that can negate the gains of both: 
 \textbf{(i) Sequential Dependency.} Semantic condensation alters the CoT’s token-level form, invalidating the patterns token-level pruners depend on and destabilizing pruning decisions. As a result, pruners trained on original CoTs may remove tokens that become essential after condensation, inducing cascading errors.
 % \textbf{(ii) Task-Agnostic Blindness.} Many existing token-level pruners~\cite{jiang2023llmlingua,pan2024llmlingua} are intentionally domain-agnostic, scoring tokens with general importance metrics rather than task-aware reasoning cues. This is especially harmful in mathematical problem solving. Tokens like numerical values, operators, and logical connectives may look redundant under generic statistics, yet are semantically indispensable for reaching the correct answer.
\textbf{(ii) Task-Agnostic Blindness.} Many token-level pruners~\cite{jiang2023llmlingua,pan2024llmlingua} are designed to be domain-agnostic and rank tokens by generic importance rather than task-specific reasoning cues. In math reasoning, this can mistakenly remove indispensable tokens such as numbers, operators, and logical connectives, leading to incorrect answers.
\textbf{(iii) Distribution Mismatch.} Token-level pruning often produces telegraphic, fragmented traces (e.g., ``calculate… then divide''), far from the fluent reasoning style used at inference. This gap both hampers step parsing and removes intermediate scaffolding, causing errors to accumulate in multi-step tasks.

To address these challenges, we design \textbf{CtrlCoT}, a dual-granularity CoT compression framework that harmonizes semantic and token-level optimizations. CtrlCoT includes three key modules to address the aforementioned challenges: 
a) Hierarchical  Reasoning Abstraction (\textbf{HRA}), 
b) Logic-Preserving Distillation (\textbf{LPD}), and 
c) Distribution-Alignment Generation (\textbf{DAG}). 
% 1.
HRA generates CoTs with different levels of semantic detail. These hierarchical traces compensate for information loss from later token pruning, thereby mitigating information loss caused by Sequential Dependency.
% 2.
LPD mitigates Task-Agnostic Blindness by distilling the pruner with logic-aware targets on mathematical CoTs, so it preserves indispensable cues (numbers, operators, connectives) across pruning ratios.
% 3.
Then, to address Distribution Mismatch, DAG uses a Multi-Ratio CoT Generator to produce fluent, coherent CoTs as supervision, aligning training traces with the inference-time reasoning style and reducing fragmentation.
Finally, we train a Budget-Controlled Reasoner (BCR) that performs controllable reasoning under a user-specified budget.
Further, to spare users from the hassle of manually setting budgets, we introduce a Budget-Free Reasoner (BFR) that automatically generates a CoT of an appropriate length.

We evaluate CtrlCoT across multiple model scales on GSM8K and MATH-500. CtrlCoT consistently improves the accuracy--cost Pareto frontier. On MATH-500 with Qwen2.5-7B-Instruct, it uses \textbf{30.7\% fewer} tokens while achieving \textbf{+7.6} accuracy over the SOTA method. On GSM8K with Qwen2.5-14B-Instruct, it cuts tokens by \textbf{55.7\%} relative to the original model with negligible accuracy loss.

In summary, our contributions are as follows:

\begin{itemize}[leftmargin=1.6em, labelsep=0.6em, itemsep=0em, topsep=2pt]
    \item To the best of our knowledge, we are the first to explore \emph{dual-granularity} CoT compression that jointly leverages semantic-level condensation and token-level pruning.
    \item We propose CtrlCoT, a framework that learns to generate high-quality compressed reasoning traces under flexible token budgets, enabling a controllable accuracy--cost trade-off.
    % \item An adaptive budget selection mechanism is designed that automatically determines the optimal compression ratio for each instance, eliminating the need for manual tuning.
    \item A budget-free model is introduced to automatically generate a CoT of appropriate length for each instance, making budget-controlled reasoning usable without hand-tuned budgets.
    \item Extensive experiments show that CtrlCoT consistently outperforms strong baselines, achieving state-of-the-art efficient reasoning.

\end{itemize}

\section{Related Work}

% lvzheqi: Efficiency Challenges in LLM Reasoning这个title不好。不像是一个领域，像是问题背景。
% zx:有的cot压缩论文会写 LLM Reasoning/Inference-time Scaling，也是背景

% Large language models have recently shown strong reasoning abilities~\citep{openai2024openaio1card,deepseekai2025deepseekr1incentivizingreasoningcapability,comanici2025gemini25pushingfrontier}, especially with chain-of-thought (CoT) prompting~\citep{NEURIPS2022_9d560961}. 
% With step-by-step rationales, they can solve multi-step math and logic problems~\citep{DBLP:conf/iclr/LuoSX0LTGLCT025,wang2025logictree} and support complex behaviors such as coding~\citep{openai2025competitiveprogramminglargereasoning} and tool- or agent-based decision making~\citep{,liu2025toolace,wang-etal-2025-x}.
% However, these gains often rely on much longer outputs: CoT substantially increases the number of generated tokens~\citep{chen2025do,fan2025missing}, leading to higher decoding latency, memory/KV-cache usage, and serving cost~\citep{qu2025survey}. 
% While system-level optimizations(better kernels~\citep{dao2023flashattention2fasterattentionbetter}, KV-cache tricks~\citep{ainslie-etal-2023-gqa}, quantization~\citep{MLSYS2024_42a452cb}, etc.) can reduce the cost per token, they typically do not reduce how many reasoning tokens are produced, motivating methods that explicitly control or compress reasoning traces.
\begin{figure*}[t]
  \vspace{-0.8cm}
  \includegraphics[width=\linewidth]{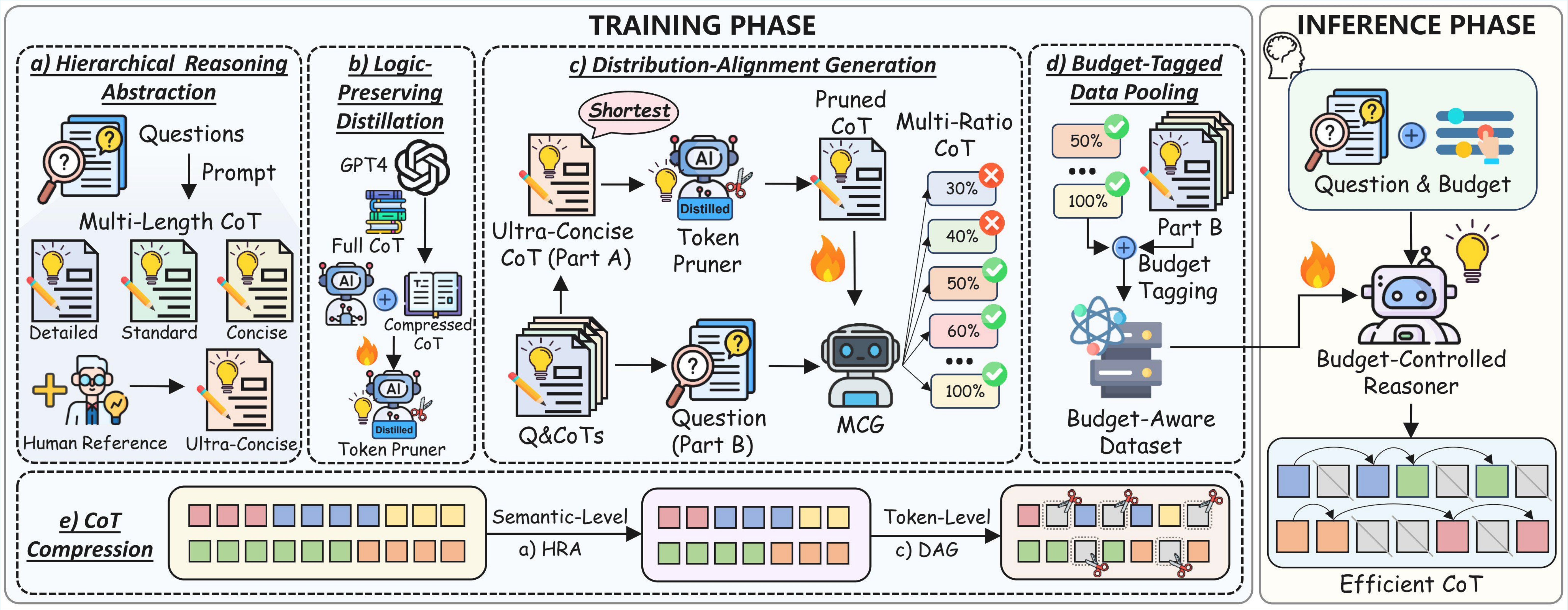}
  \vspace{-0.2cm}
\caption{
Framework of CtrlCoT. In the training stage, HRA first performs semantic-level compression, followed by token-level compression with LPD and DAG; the resulting CoTs are then aggregated for data pooling and model training. In the inference stage, the LLM performs efficient budget-conditioned reasoning given a user-specified token budget.
}
\vspace{-0.2cm}
\end{figure*}

\noindent\textbf{Efficient Reasoning for LLMs.}
\label{sec:rw_budget}
Although CoT prompting can substantially improve reasoning quality, it also introduces a pronounced efficiency challenge.
There has been research on Efficient Learning for LLMs from perspectives such as parameter quantization~\citep{lin2024duquant}, collaborative inference~\citep{lv2025collaboration}, and so on.
CoT substantially increases the number of generated tokens~\citep{chen2025do,fan2025missing}, leading to higher decoding latency~\citep{leviathan2023fast}, memory usage~\citep{liu2024minicache}, and serving cost~\citep{qu2025survey}.
These overheads directly reduce throughput and increase inference budgets, making large-scale deployment and latency-sensitive applications substantially harder.
While system-level optimizations such as better kernels~\citep{dao2023flashattention2fasterattentionbetter}, KV-cache techniques~\citep{ainslie-etal-2023-gqa}, and quantization~\citep{frantar2023optq,MLSYS2024_42a452cb} can reduce the cost per token, they typically do not reduce the number of reasoning tokens produced. This motivates methods that explicitly control or compress reasoning traces~\citep{alomrani2025reasoning}.

% 大型语言模型最近展现出了强大的推理能力，尤其是在与链式思维（CoT）提示相结合时。通过明确的逐步推理过程，LLM 能够解决多步骤的数学和逻辑问题，生成和调试代码，并在基于工具和代理的流程中做出更可靠的决策。然而，这些推理能力的提升通常伴随着更长的输出。CoT 通常会将简短的答案扩展为数十甚至数百个中间标记，这增加了自回归解码步骤，扩大了 KV 缓存，并放大了延迟、内存使用和服务器成本。现有的系统级优化（更好的内核、KV 缓存技巧、量化、推测性解码等）可以降低固定解码工作负载的成本，但它们很少改变模型生成的推理文本量。因此，效率瓶颈越来越多地来自 CoT 本身标记的低效性，这促使人们采用能够明确控制或减少推理痕迹长度的方法

% lvzheqi：可以考虑不要只用同一种字体，可以考虑更吸引人的字体用在重要的部分比如重要组件用comic，非重要部分用相对低调的字体比如Times。这里的TRAINING和INFERENCE啥的就很好，跟图中其他部分字体区分开了。
% zx:两个压缩的地方换成了别的字体

\noindent\textbf{CoT Compression.}
% 无需训练：prompt，解码，截断
% 训练：SFT（隐式推理）和RL
CoT compression reduces generated tokens to improve reasoning efficiency.
Following the survey taxonomy~\citep{sui2025stop}, we categorize prior work by the \emph{intervention locus} in the reasoning pipeline: \emph{model-}, \emph{output-}, and \emph{input-based} methods.
This taxonomy is orthogonal to the semantic-/token-level view, which describes \emph{what} is compressed rather than \emph{where} it is applied.
% CoT compression reduces generated tokens to improve reasoning efficiency, and can be grouped into three categories~\citep{sui2025stop}.
\emph{Model-based} methods explicitly optimize the model to produce shorter CoTs, typically via supervised fine-tuning (SFT) on variable-length CoT data~\citep{xia2025tokenskip,DBLP:conf/aaai/KangSCZ25,ma2025cotvalvelengthcompressiblechainofthoughttuning} or reinforcement learning (RL) with length reward~\citep{luo2025o1,aggarwal2025l1,team2025kimi}.
% \emph{Output-based} methods focus on modifying the reasoning outputs to encourage concise reasoning: some instantiate a \emph{dynamic reasoning paradigm} that adaptively adjusts reasoning depth based on reward signals~\citep{sun2024fast}, confidence~\citep{ding-etal-2025-dynamic}, or consistency~\citep{wang2025sampling}, while others compress CoTs into more compact latent representations to reduce decoding cost~\citep{hao2025training,shen-etal-2025-codi,cheng2024compressed}.
\emph{Output-based} methods modify reasoning outputs to promote conciseness: some adapt reasoning depth using reward~\citep{sun2024fast}, confidence~\citep{ding-etal-2025-dynamic}, or consistency~\citep{wang2025sampling}, while others compress CoTs into compact latent representations to reduce decoding cost~\citep{hao2025training,shen-etal-2025-codi,cheng2024compressed}.
Different from the previous two categories, \emph{input-based} methods enforce length constraints or route the LLM based on the input prompts.
Our method is \emph{model-based} (SFT) and is orthogonal to output- and input-based strategies, making it complementary to them. 
% For example, prompt-based routing decides the model size to use for each instance, while our model controls how to express reasoning under a given budget.
\section{Method}
\label{sec:method}
% lzq:方法章节现在小标题太多了，training and inference procedure可以单拎出来一个子章节。3.3 3.4应该是大块，但是考虑到这种工作已经很多，你直接起这么大的标题也不好。Semantic-Level Compression和token-level compression都可以换个名字。比如global-aware/local-aware ...(这个名字也不好，意思就是换个名字)
% \subsection{Problem Setup and Budget-Conditioned Reasoning}
\subsection{Problem Setup and Overview}
\label{sec:setup}

We study budget-conditioned math reasoning, where a model takes a problem $x$ and a user-specified budget $b$, and outputs a final answer $y$ together with a reasoning trace $r$ whose length is controlled by $b$.
Our goal is to learn a conditional generator:

\begin{equation}
p_{\theta}(r, y \mid x, b),
\end{equation}
so that the model remains accurate while adjusting the verbosity of $r$ to the given budget.

% Our approach defines a two-axis compression space and collects CoTs along two complementary dimensions: (i) Hierarchical Reasoning Abstraction, which semantically compresses CoTs by varying abstraction levels and step granularity via prompting; and (ii) Multi-Ratio Trace Pruning, which token-level prunes CoTs using a learned pruner under controllable ratios. We pool CoTs from both axes and conduct budget-conditioned training with budget-tagged supervision to obtain a Budget-Controlled Reasoner.

Our approach consists of three modules: Hierarchical Reasoning Abstractor (HRA), Logic-Preserving Distillator (LPD), and Distribution-Alignment Generator (DAG). HRA produces multi-level semantic traces, LPD preserves key logical cues across pruning ratios, and DAG generates fluent supervision to reduce distribution mismatch. Finally, we train two reasoners for budget-controlled and budget-free inference, respectively.

% \subsection{CtrlCoT}
\subsection{Hierarchical Reasoning Abstraction}
% \label{sec:semantic}
% To obtain semantic-level compressed CoTs, we construct a set of four CoT verbosity tiers for each instance:
% Detailed CoT, Standard CoT, Concise CoT, and Ultra-Concise CoT.
% These tiers differ in both verbosity and step granularity, and together form discrete length points along the semantic compression axis.
\label{sec:semantic}
In this module, we create four semantically compressed CoT tiers for each instance: Detailed, Standard, Concise, and Ultra-Concise.
They differ in verbosity and step granularity, yielding discrete length points in the abstraction space.

% \paragraph{Hierarchical Abstraction.}
% For each instance $(x,y)\in\mathcal{D}$, we design three prompt templates
% $\{\pi_{\text{v}}, \pi_{\text{s}}, \pi_{\text{c}}\}$ corresponding to Detailed, Standard, and Concise styles, respectively.
% Each template instructs the model to generate a CoT and an answer, while controlling the CoT verbosity as requested.
% Crucially, the model is given the problem $x$, the ground-truth answer $y$, and the selected template, and is asked to generate a supporting CoT:
\paragraph{Base-Tier Abstraction.}
% For each instance $(x,y)\in\mathcal{D}$, we design three abstraction-tier prompt templates
% $\{\pi_{\text{d}}, \pi_{\text{s}}, \pi_{\text{c}}\}$ corresponding to \textit{Detailed}, \textit{Standard}, and \textit{Concise} CoTs.
% These templates control the abstraction level (verbosity and step granularity) while eliciting a CoT and the final answer.
% Concretely, given the problem $x$, the ground-truth answer $y$, and a selected abstraction template, the model is instructed to generate a supporting CoT:

For each instance $(x, y) \in \mathcal{D}$, where $\mathcal{D}$ denotes the training set, $x$ is the input problem and $y$ is the ground-truth answer, we design three abstraction-tier prompt templates ${\pi_{\text{d}}, \pi_{\text{s}}, \pi_{\text{c}}}$ corresponding to \textit{Detailed}, \textit{Standard}, and \textit{Concise} CoTs.
These templates control the abstraction level (verbosity and step granularity) while eliciting a supporting CoT along with the final answer. Concretely, given $x$, $y$, and a selected template, the model is instructed to generate a CoT that justifies $y$:

% lvzheqi:全文加粗的部分太多了，注意一下，跟打补丁一样。比如problem和ground-truth answer不用加粗。同上，这种乱加粗的地方一眼LLM写的，需要check一下。Introduction里其实我觉得加粗也比较多，但还能忍受。到这里这个加粗就有点过于泛滥了。
% zx:已修改
\begin{equation}
r^{\text{hra}}_{t} \sim p(\cdot \mid x, y, \pi_{t}), \quad t \in \{\text{d}, \text{s}, \text{c}\}.
\end{equation}

% \paragraph{Maximal Abstraction.}
% To obtain an even shorter Ultra-Concise CoT, prompting alone is often insufficient to reliably reach the desired minimal length without losing essential reasoning.
% We therefore additionally provide a human-written, extremely concise reference CoT $r^{\text{ref}}$ and ask the model to produce a brief restatement that preserves the key logic.
% Formally,
% \begin{equation}
% r^{\text{hra}}_{\text{u}} \sim p(\cdot \mid x, y, r^{\text{ref}}, \pi_{\text{u}}),
% \end{equation}
% where $\pi_{\text{u}}$ is a template that explicitly instructs concise rephrasing with minimal tokens.

\paragraph{Reference-Guided Minimal Abstraction.}
To obtain the \textit{Ultra-Concise} CoT, abstraction-tier prompting alone is often insufficient to consistently reach a minimal length without dropping key logic.
We therefore provide a human-written concise reference trace $r^{\text{ref}}$ and ask the model to produce a minimal CoT that preserves the core reasoning.
Formally,
\begin{equation}
r^{\text{hra}}_{\text{u}} \sim p(\cdot \mid x, y, r^{\text{ref}}, \pi_{\text{u}}),
\end{equation}
where $\pi_{\text{u}}$ is an abstraction template that enforces a minimal-token realization.

\paragraph{Answer-Consistency Filtering.}
Since generated CoTs may still end with an incorrect answer, we apply an answer-consistency filter for each verbosity tier.
For a generated trace $r^{\text{hra}}_{t}$, we parse the predicted final answer $\hat{y}_{t}$ from the model output and retain the trace only if it matches the ground-truth answer $y$:
\begin{equation}
\mathcal{R}^{\text{hra}}_{\text{correct}}(x) =
\left\{ r^{\text{hra}}_{t} \,\middle|\, \hat{y}_{t}=y, \; t\in\{\text{d},\text{s},\text{c},\text{u}\} \right\}.
\end{equation}
% Through this process, we obtain a set of correctness-filtered CoTs with varying degrees of semantic compression, serving as the semantic axis of our dual-granularity compression space.
% This yields correctness-filtered CoTs with different degrees of semantic compression, forming the Hierarchical Reasoning Abstraction axis in our dual-granularity compression space.
This completes the Hierarchical Reasoning Abstraction by producing CoTs at multiple abstraction levels, forming the semantic axis of our dual-granularity compression space.

\subsection{Logic-Preserving Distillation}

% To enable token-level compression in CtrlCoT, a token pruner is required to shorten CoTs under a controllable compression ratio.
After semantic compression with HRA, CtrlCoT applies token-level compression to further remove token redundancy under a controllable ratio, which requires a token pruner to shorten CoTs accordingly.
We use LLMLingua2~\citep{pan2024llmlingua} as an off-the-shelf token-level pruner: given a text sequence and a target compression ratio, it removes tokens deemed unimportant and outputs a pruned text whose length matches the specified ratio while aiming to preserve the original content.
Formally, for a CoT $r^{\text{u}}(x)$ and a compression ratio $\gamma$, the pruned variant is
\begin{equation}
\tilde{r}^{\gamma}(x) = \mathcal{P}\!\left(r^{\text{u}}(x);\gamma\right),
\end{equation}
where $\mathcal{P}$ denotes the pruner and $\gamma$ controls the target compression strength.
However, as a general-purpose pruner, LLMLingua2 can exhibit task-agnostic blindness on math CoTs and discard logic-critical tokens such as numbers, operators, and intermediate expressions.

To address this, we perform Logic-Preserving Distillation on a small set of math CoTs: we use GPT-4~\citep{openai2024gpt4technicalreport} to generate high-quality token-pruned targets and fine-tune LLMLingua2 on these (CoT, pruned-CoT) pairs.\footnote{See Appendix~\ref{sec:distill_pruner} for details.} This improves pruning accuracy on math reasoning text, encouraging the pruner to retain logic-bearing cues under different pruning ratios, and yields more reliable multi-strength variants $\tilde{r}^{\gamma}(x)$ for subsequent training.

\subsection{Distribution-Alignment Generation}
\label{sec:dag}

To make token pruning more accurate and mitigate the distribution mismatch introduced by pruned, telegraphic CoTs, we train a Multi-Ratio CoT Generator (MCG) to produce fluent and coherent CoTs under controllable ratios.
We split $\mathcal{D}$ into disjoint $\mathcal{D}_A$ and $\mathcal{D}_B$.
MCG is trained on $\mathcal{D}_A$ and then run at multiple compression strengths on $\mathcal{D}_B$ to obtain ratio-controlled, distribution-aligned CoTs. This training procedure is inspired by ~\citep{xia2025tokenskip}.

\paragraph{MCG Training.}
\label{sec:dag_train}
% For each instance $(x,y)\in\mathcal{D}_A$, we select the shortest correct CoT along the \emph{Reasoning Abstraction} axis as the seed trace.% In our construction, this corresponds to the \textit{Ultra-Concise} CoT obtained in Section~\ref{sec:semantic}.

For each instance $(x,y)\in\mathcal{D}_A$, we use the \textit{Ultra-Concise} correct CoT obtained in Section~\ref{sec:semantic} as the seed trace, since it is the shortest among the abstraction tiers.
We denote the selected seed as $r^{\text{u}}(x)$.

We then construct multi-ratio targets by applying the token pruner $\mathcal{P}$ to $r^{\text{u}}(x)$ at multiple ratios $\gamma\in\Gamma$,
\begin{equation}
\tilde{r}^{\gamma}(x) = \mathcal{P}\!\left(r^{\text{u}}(x);\gamma\right), \quad \gamma\in\Gamma,
\end{equation}
where $\Gamma$ is a predefined set of compression ratios (we use $\Gamma=\{0.3,0.4,\ldots,1.0\}$).
We pair each training instance with a prompt that explicitly specifies $\gamma$ (and enforces a coherent reasoning style), and fine-tune an LLM to learn ratio-controlled generation:
\begin{equation}
p_{\phi}\!\left(r, y \mid x, \gamma \right).
\end{equation}
Although the pruned targets $\tilde{r}^{\gamma}(x)$ may be fragmentary, the autoregressive generator learns to realize them as fluent and step-coherent traces under the same ratio constraint, thereby aligning the training distribution with inference-time reasoning.

\paragraph{Distribution-Aligned CoT Generation.}
\label{sec:dag_apply}
After training, we apply MCG to $\mathcal{D}_B$ to obtain multiple distribution-aligned CoTs per problem.
Given $(x,y)\in\mathcal{D}_B$, we run MCG with different ratios to generate CoTs of varying lengths:
\begin{equation}
r^{\text{dag}}_{\gamma} \sim p_{\phi}(\cdot \mid x, \gamma), \quad \gamma \in \Gamma.
\end{equation}
Finally, to ensure correctness, we apply the same answer-consistency filtering as in Section~\ref{sec:semantic}, retaining only generations whose final answer matches the ground-truth $y$.
% The resulting set $\{r^{\text{dag}}_{\gamma}\}_{\gamma\in\Gamma_B}$ forms the DAG axis of our dual-granularity compression space, providing a dense spectrum of fluent, ratio-controlled CoTs for subsequent training.
Since the training data is token-pruned on top of semantic abstraction, the resulting traces $\{r^{\text{dag}}_{\gamma}\}_{\gamma\in\Gamma_B}$ naturally combine semantic- and token-level compression.

\subsection{Reasoner Training and Inference}
After obtaining all CoTs, we introduce the Budget-Controlled Reasoner and the Budget-Free Reasoner.
\paragraph{Budget-Tagged Data Pooling.}
% After generating CoTs along both axes, we aggregate them for each problem $x$ into a unified pool:
We first aggregate the CoTs generated along both axes for each problem $x$ into a unified pool:
\begin{equation}
\mathcal{R}(x) =
\underbrace{\{r^{\text{hra}}_{k}\}_{k=1}^{K}}_{\text{semantic-level}}
\;\cup\;
\underbrace{\{r^{\text{dag}}_{\gamma}\}_{\gamma\in\Gamma}}_{\text{semantic+token}}.
\end{equation}

For each trace $r \in \mathcal{R}(x)$, we compute its CoT length in tokens, denoted as
\begin{equation}
b(r)=|r|.
\end{equation}
We then convert each $(x,y,r)$ into a budget-tagged instruction by explicitly requesting an approximate token usage in the prompt, e.g., ``\textit{Please answer using approximately $b(r)$ tokens for the reasoning}''.
This results in a budget-aware dataset where each problem is paired with multiple CoTs of different lengths and compression styles, each associated with its own budget label.

\begin{figure}[t]       % [t]=顶部 [h]=当前位置 [b]=底部 [!ht]=尽量当前位置
  \centering
  \includegraphics[width=\linewidth]{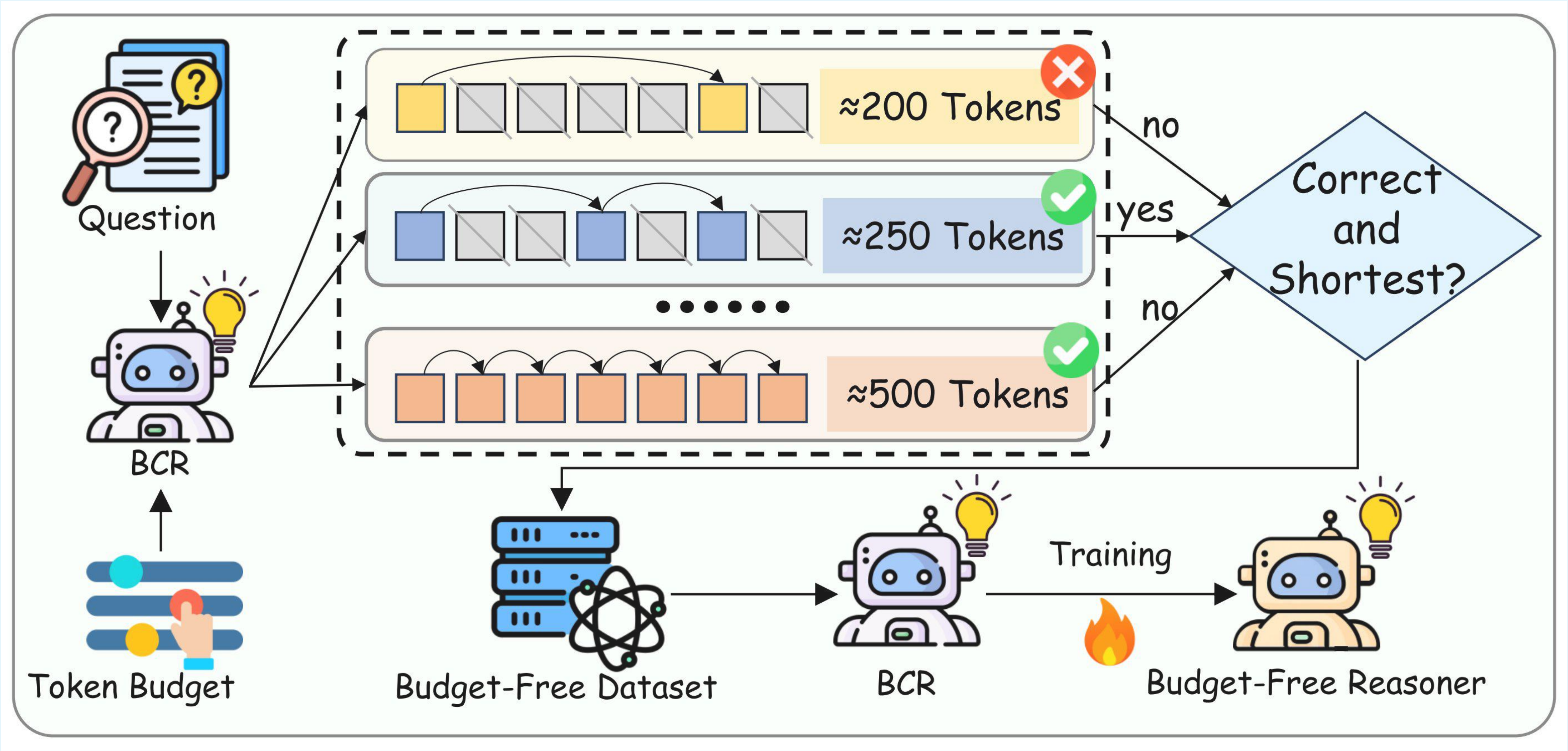} % 也可以写成 width=0.8\linewidth
  \vspace{-0.5cm}
  \caption{The process of constructing a Budget-Free Reasoner. By training on the shortest correct CoTs, BFR enables efficient automatic reasoning.
}
  \label{fig:auto_budget}
\vspace{-0.3cm}
\end{figure}
\label{sec:reasoner}

\paragraph{Budget-Controlled Reasoner (BCR).}
We fine-tune the language model with standard supervised fine-tuning (SFT) on budget-tagged examples:
\begin{equation}
\mathcal{L}_{\text{SFT}} = -\log p_{\theta}\!\left(r, y \mid x, b(r)\right),
\end{equation}
where $b(r)$ is the token budget tag paired with CoT $r$.
This training process yields a Budget-Controlled Reasoner (BCR) that performs controllable reasoning under a user-specified budget.

At test time, we feed BCR the problem $x$ together with a user-specified token budget $b$ and generate
\begin{equation}
(r, y) \sim p_{\theta}(\cdot \mid x, b).
\end{equation}
As a result, BCR follows the requested budget for length control while integrating semantic conciseness with token-level skipping.

\paragraph{Budget-Free Reasoner (BFR).}
While BCR requires a user-specified budget, we further train a Budget-Free Reasoner (BFR) that automatically produces a near-minimal correct CoT.
This stage is performed on the $\mathcal{D}_A$ split. For each instance $(x,y)\in\mathcal{D}_A$, we run BCR with a set of candidate budgets $\mathcal{B}$, filter generations by answer consistency, and select the shortest correct CoT among candidate budgets:
\begin{equation}
r^{\dagger}(x)=\arg\min_{b\in\mathcal{B}:\,\hat{y}_b=y} |r_b|, 
\quad (r_b,\hat{y}_b)\sim p_{\theta}(\cdot\mid x,b).
\end{equation}
We then construct $\mathcal{D}_{\text{BFR}}=\{(x,r^{\dagger}(x),y)\}$ and perform standard SFT to obtain a model that directly generates concise, correctness-preserving CoTs.

\begin{table*}[t]
\centering
\small
\setlength{\tabcolsep}{7pt}
\renewcommand{\arraystretch}{0.93}
\resizebox{\linewidth}{!}{
\begin{tabular}{l l l | r r r r | r r r r}
\toprule[1.5pt]
% \multirow{2}{*}{\multicolumn{1}{c}{\textbf{Model}}} &
\multirow{2}{*}{\textbf{Model}} &
% \multirow{2}{*}{\multicolumn{1}{c}{\textbf{CS.}}} &
% \multirow{2}{*}{\multicolumn{1}{c}{\textbf{Methods}}} &
\multirow{2}{*}{\textbf{CS.}} &
\multirow{2}{*}{\textbf{Methods}} &
\multicolumn{4}{c|}{\textbf{GSM8K}} &
\multicolumn{4}{c}{\textbf{MATH-500}} \\
\cmidrule(lr){4-7}\cmidrule(lr){8-11}
& & &
\textbf{Acc. $\uparrow$} &
\textbf{Tokens $\downarrow$} &
\textbf{CR. $\downarrow$} &
\textbf{TE. $\uparrow$} &
\textbf{Acc. $\uparrow$} &
\textbf{Tokens $\downarrow$} &
\textbf{CR. $\downarrow$} &
\textbf{TE. $\uparrow$} \\
\midrule

% ================= Qwen2.5-3B =================
\multirow{13}{*}{\textbf{Qwen2.5-3B}}
& --- & Original &
83.24 & 316.94 & 1.00 & 3.81 &
63.20 & 575.66 & 1.00 & 9.11 \\
\cmidrule(lr){2-11}

& \multirow{4}{*}{High} & \textcolor{black!60}{LC\text{-}Prompt} &
\textcolor{black!60}{83.32} & \textcolor{black!60}{279.85} & \textcolor{black!60}{0.88} & \textcolor{black!60}{29.77} &
\textcolor{black!60}{61.40} & \textcolor{black!60}{534.98} & \textcolor{black!60}{0.93} & \textcolor{black!60}{11.48} \\
&  & Truncation &
36.32 & 244.22 & 0.77 & 14.87 &
41.20 & 437.92 & 0.76 & 9.41 \\
&  & TokenSkip &
71.65 & 157.11 & 0.50 & 45.60 &
39.80 & 336.75 & 0.58 & 11.82 \\
&  & \textbf{Ours} &
\textbf{75.59} & \textbf{112.76} & \textbf{0.36} & \textbf{67.03} &
\textbf{46.80} & \textbf{315.49} & \textbf{0.55} & \textbf{14.83} \\
\cmidrule(lr){2-11}

& \multirow{4}{*}{Mid} & \textcolor{black!70}{LC\text{-}Prompt} &
\textcolor{black!60}{82.79} & \textcolor{black!60}{296.34} & \textcolor{black!60}{0.94} & \textcolor{black!60}{27.94} &
\textcolor{black!60}{61.60} & \textcolor{black!60}{564.82} & \textcolor{black!60}{0.98} & \textcolor{black!60}{10.91} \\
&  & Truncation &
54.21 & 276.08 & 0.87 & 19.63 &
50.20 & 483.86 & 0.84 & 10.37 \\
&  & TokenSkip &
75.74 & 186.18 & 0.59 & 40.68 &
49.00 & 399.20 & 0.69 & 12.27 \\
&  & \textbf{Ours} &
\textbf{77.10} & \textbf{125.44} & \textbf{0.40} & \textbf{61.47} &
\textbf{51.20} & \textbf{356.06} & \textbf{0.62} & \textbf{14.38} \\
\cmidrule(lr){2-11}

& \multirow{4}{*}{Low} & \textcolor{black!60}{LC\text{-}Prompt} &
\textcolor{black!60}{82.87} & \textcolor{black!60}{294.62} & \textcolor{black!60}{0.93} & \textcolor{black!60}{28.13} &
\textcolor{black!60}{62.00} & \textcolor{black!60}{561.29} & \textcolor{black!60}{0.98} & \textcolor{black!60}{11.05} \\
&  & Truncation &
69.29 & 296.88 & 0.94 & 23.34 &
\textbf{55.00} & 518.07 & 0.90 & 10.62 \\
&  & TokenSkip &
\textbf{79.98} & 230.42 & 0.73 & 34.71 &
53.80 & 444.36 & 0.77 & 12.11 \\
&  & \textbf{Ours} &
79.83 & \textbf{193.43} & \textbf{0.61} & \textbf{41.27} &
54.40 & \textbf{403.12} & \textbf{0.70} & \textbf{13.49} \\

\midrule

% ================= Qwen2.5-7B =================
\multirow{13}{*}{\textbf{Qwen2.5-7B}}
& --- & Original &
91.58 & 299.22 & 1.00 & 3.27 &
71.20 & 574.58 & 1.00 & 7.70 \\
\cmidrule(lr){2-11}

& \multirow{4}{*}{High} & \textcolor{black!60}{LC\text{-}Prompt} &
\textcolor{black!60}{89.84} & \textcolor{black!60}{214.85} & \textcolor{black!60}{0.72} & \textcolor{black!60}{41.81} &
\textcolor{black!60}{71.00} & \textcolor{black!60}{514.08} & \textcolor{black!60}{0.89} & \textcolor{black!60}{13.81} \\
&  & Truncation &
45.64 & 240.98 & 0.81 & 18.94 &
46.00 & 437.80 & 0.76 & 10.51 \\
&  & TokenSkip &
83.40 & 149.43 & 0.50 & 55.81 &
50.40 & 325.68 & 0.57 & 15.48 \\
&  & \textbf{Ours} &
\textbf{85.82} & \textbf{138.41} & \textbf{0.46} & \textbf{62.01} &
\textbf{58.00} & \textbf{225.61} & \textbf{0.39} & \textbf{25.71} \\
\cmidrule(lr){2-11}

& \multirow{4}{*}{Mid} & \textcolor{black!60}{LC\text{-}Prompt} &
\textcolor{black!60}{90.37} & \textcolor{black!60}{238.85} & \textcolor{black!60}{0.80} & \textcolor{black!60}{37.84} &
\textcolor{black!60}{70.60} & \textcolor{black!60}{528.88} & \textcolor{black!60}{0.92} & \textcolor{black!60}{13.35} \\
&  & Truncation &
64.75 & 267.78 & 0.89 & 24.18 &
55.60 & 482.67 & 0.84 & 11.52 \\
&  & TokenSkip &
86.35 & 171.70 & \textbf{0.57} & 50.29 &
58.40 & 376.05 & 0.65 & 15.53 \\
&  & \textbf{Ours} &
\textbf{87.72} & \textbf{170.74} & \textbf{0.57} & \textbf{51.37} &
\textbf{60.40} & \textbf{267.07} & \textbf{0.46} & \textbf{22.62} \\
\cmidrule(lr){2-11}

& \multirow{4}{*}{Low} & \textcolor{black!60}{LC\text{-}Prompt} &
\textcolor{black!60}{89.92} & \textcolor{black!60}{243.41} & \textcolor{black!60}{0.81} & \textcolor{black!60}{36.94} &
\textcolor{black!60}{70.20} & \textcolor{black!60}{533.52} & \textcolor{black!60}{0.93} & \textcolor{black!60}{13.16} \\
&  & Truncation &
78.47 & 283.86 & 0.95 & 27.64 &
61.20 & 518.06 & 0.90 & 11.81 \\
&  & TokenSkip &
88.63 & 209.33 & 0.70 & 42.34 &
64.20 & 436.93 & 0.76 & 14.69 \\
&  & \textbf{Ours} &
\textbf{89.31} & \textbf{198.11} & \textbf{0.66} & \textbf{45.08} &
\textbf{68.80} & \textbf{428.68} & \textbf{0.75} & \textbf{16.05} \\

\midrule

\multirow{13}{*}{\textbf{Qwen2.5-14B}}
& --- & Original
& 93.03 & 313.94 & 1.00 & 29.63
& 75.80 & 583.66 & 1.00 & 12.99 \\
\cmidrule(lr){2-11}

% ----- High (0.5) -----
& \multirow{4}{*}{High} & \textcolor{black!60}{LC\text{-}Prompt}
& \textcolor{black!60}{94.39} & \textcolor{black!60}{246.98} & \textcolor{black!60}{0.79} & \textcolor{black!60}{38.22}
& \textcolor{black!60}{76.60} & \textcolor{black!60}{524.37} & \textcolor{black!60}{0.90} & \textcolor{black!60}{14.61} \\
&  & Truncation
& 38.67 & 244.96 & 0.78 & 15.78
& 46.80 & 443.17 & 0.76 & 10.56 \\
&  & TokenSkip
& 90.37 & 158.44 & 0.50 & 57.04
& 59.40 & 346.04 & 0.59 & 17.17 \\
&  & \textbf{Ours}
& \textbf{90.67} & \textbf{139.09} & \textbf{0.44} & \textbf{65.19}
& \textbf{66.60} & \textbf{270.80} & \textbf{0.46} & \textbf{24.59} \\
\cmidrule(lr){2-11}

% ----- Mid (0.6) -----
& \multirow{4}{*}{Mid} & \textcolor{black!60}{LC\text{-}Prompt}
& \textcolor{black!60}{93.78} & \textcolor{black!60}{276.52} & \textcolor{black!60}{0.88} & \textcolor{black!60}{33.92}
& \textcolor{black!60}{74.20} & \textcolor{black!60}{549.37} & \textcolor{black!60}{0.94} & \textcolor{black!60}{13.51} \\
&  & Truncation
& 60.80 & 275.82 & 0.88 & 22.04
& 56.80 & 489.82 & 0.84 & 11.60 \\
&  & TokenSkip
& 91.89 & 191.24 & 0.61 & 48.05
& 63.80 & 398.32 & 0.68 & 16.02 \\
&  & \textbf{Ours}
& \textbf{92.04} & \textbf{185.41} & \textbf{0.59} & \textbf{49.64}
& \textbf{69.20} & \textbf{318.73} & \textbf{0.55} & \textbf{21.71} \\
\cmidrule(lr){2-11}

% ----- Low (0.7) -----
& \multirow{4}{*}{Low} & \textcolor{black!60}{LC\text{-}Prompt}
& \textcolor{black!60}{93.33} & \textcolor{black!60}{285.60} & \textcolor{black!60}{0.91} & \textcolor{black!60}{32.68}
& \textcolor{black!60}{75.80} & \textcolor{black!60}{551.17} & \textcolor{black!60}{0.94} & \textcolor{black!60}{13.75} \\
&  & Truncation
& 76.19 & 295.15 & 0.94 & 25.82
& 63.00 & 525.75 & 0.90 & 11.98 \\
&  & TokenSkip
& \textbf{92.95} & 224.26 & 0.71 & 41.45
& 71.00 & 441.72 & 0.76 & 16.07 \\
&  & \textbf{Ours}
& 92.72 & \textbf{210.24} & \textbf{0.67} & \textbf{44.10}
& \textbf{72.40} & \textbf{408.27} & \textbf{0.70} & \textbf{17.73} \\

\bottomrule[1.5pt]
\end{tabular}
}
\vspace{-0.2cm}
\caption{Performance comparison between Ours and other methods on Qwen2.5-3B/7B/14B-Instruct. Since LC-Prompt offers only mild compression, later analysis focuses on Truncation and TokenSkip for fair comparison. Under almost all settings, our method achieves the highest accuracy with the fewest tokens, resulting in the best token efficiency.}
\label{tab:main_table}
\vspace{-0.3cm}
\end{table*}

\section{Experiments}

\subsection{Experimental Setup}

\paragraph{Datasets and Metrics.}
We conduct experiments on two widely-used mathematical reasoning benchmarks: GSM8K~\citep{cobbe2021trainingverifierssolvemath} and MATH~\citep{DBLP:conf/nips/HendrycksBKABTS21}. 
To keep evaluation efficient while remaining comparable to prior studies~\citep{DBLP:conf/iclr/LightmanKBEBLLS24}, we report results on the MATH-500 split of MATH.
We report (i) Accuracy, (ii) CoT Token Count, and (iii) Compression Ratio (CR) relative to the uncompressed reference.
All evaluations are performed with the official DeepSeekMath evaluation scripts~\citep{deepseek-math} for consistent scoring.
% lzq：这种bad ending减少一点，对观感影响比较大，也就是说下一行就几个字母/一个单词的这种。
% 已解决

% To explicitly capture the accuracy--cost trade-off, we further report Token Efficiency (TE), which normalizes accuracy by the average number of generated CoT tokens:
% \begin{equation}
% \mathrm{TE} = \frac{\mathrm{Acc}}{\mathrm{Tokens}} \times 100,
% \end{equation}
% Higher TE indicates that the model achieves stronger performance per generated token, reflecting more efficient use of the output budget.

To capture the accuracy--cost trade-off, we report Token Efficiency (TE), defined as accuracy normalized by the average CoT tokens:
\begin{equation}
\mathrm{TE} = \frac{\mathrm{Acc}}{\mathrm{Tokens}} \times 100.
\end{equation}
Higher TE indicates better performance per generated token, reflecting more efficient use of the output budget.

% \paragraph{Models and Datasets.}
% We evalur method using the Qwen2.5-Instruct models of three scales (3B, 7B, and 14B). 
% Experiments are conducted on two math reasoning benchmarks: GSM8K~\citep{cobbe2021gsm8k} and MATH-500~\citep{hendrycks2021math}. 
% Each dataset is used independently for training and evaluation.

\paragraph{Baselines.}
% We benchmark CtrlCoT against three representative approaches:
% (1) \textbf{Prompt-based length control.} We include length-conditioning prompt (LC-Prompt), a semantic-level CoT compression approach that asks the model to reduce its reasoning length by a specified proportion during generation;
% (2) \textbf{Truncation.} As a simple reference point, we cap the maximum CoT length at a target ratio, yielding a brute-force compression baseline; and
% (3) \textbf{TokenSkip}~\citep{xia2025tokenskip}, a representative token-level CoT compression method that enables controllable shortening via token skipping.
% We denote different compression intensities using Compression Strength (CS), where \textbf{High} indicates more aggressive compression and \textbf{Low} indicates milder compression.
We benchmark CtrlCoT against three baselines: (1) \textbf{LC-Prompt}, a prompt-based method that conditions generation on a target length reduction; (2) \textbf{Truncation}, which caps the CoT length at a target ratio; and (3) \textbf{TokenSkip}~\citep{xia2025tokenskip}, which performs token-level shortening via skipping. We report results under different Compression Strength (CS) settings, where \textbf{High} is more aggressive and \textbf{Low} is milder.

\paragraph{Implementation Details.}
Our experiments are implemented with instruction-tuned backbones from two model families: Qwen2.5-Instruct at three scales (3B, 7B, and 14B)~\citep{yang2024qwen2technicalreport}, and LLaMA-3.1-8B-Instruct~\citep{grattafiori2024llama3herdmodels}.
We perform parameter-efficient fine-tuning with LoRA~\citep{hu2022lora}, using the LlamaFactory toolkit~\citep{zheng2024llamafactory} for training. Experiments are conducted on NVIDIA RTX 4090D GPUs, except for Qwen2.5-14B-Instruct, which is trained on NVIDIA A800 GPUs.

% Unless stated otherwise, we report results on Qwen2.5-7B trained on MATH, which serves as our default setting.

\subsection{Main Results}
Table~\ref{tab:main_table} compares Ours with baselines on Qwen2.5-Instruct models (3B/7B/14B) under multiple compression settings. LC-Prompt preserves accuracy but provides limited compression, so it is excluded from later comparisons at matched CoT lengths. Truncation reduces tokens but consistently hurts accuracy, while TokenSkip compresses tokens with smaller losses. Our method further improves this trade-off, using fewer tokens while achieving higher accuracy.

Concretely, on GSM8K, our approach consistently surpasses TokenSkip across different budgets. For example, on Qwen2.5-3B-Instruct under high compression strength, our method improves accuracy from 71.65\% to 75.59\% while reducing the token count from 157.11 to 112.76 compared with TokenSkip. Similar trends hold under lower compression strength: at low compression strength, our method reduces the token count by 16\% while achieving accuracy comparable to TokenSkip. The same pattern extends to larger models: \textit{Ours} continues to deliver higher accuracy at lower cost on both 7B and 14B, indicating stable gains across scales.

% On the more challenging MATH-500 benchmark, the advantage becomes even more pronounced. On the 3B model, across different compression strengths, our method consistently uses 20--40 fewer tokens than TokenSkip while achieving higher accuracy; under \textbf{Low} compression strength, the absolute accuracy gain reaches up to 7 percentage points. The gains are larger with the 7B model. For instance, under high compression strength, our approach reaches 58.00\% accuracy with only 225.61 tokens on Qwen2.5-7B-Instruct, compared to TokenSkip's 50.40\% with 325.68 tokens. This trend persists on Qwen2.5-14B-Instruct, where \textit{Ours} achieves the highest token efficiency: on GSM8K it reduces tokens by up to 56\% relative to the original model with negligible accuracy drop, and on MATH-500 (high compression) it uses a further 22\% fewer tokens than TokenSkip while improving accuracy by 7.2 points.

On the harder MATH-500 benchmark, the gap widens: with the 3B model, across compression strengths, our method uses 20–40 fewer tokens than TokenSkip while achieving higher accuracy, with up to +7.0 points under high compression. The gains are larger at 7B and 14B—for instance, under High compression on Qwen2.5-7B-Instruct, we reach 58.00\% with 225.61 tokens vs. TokenSkip’s 50.40\% with 325.68 tokens; on Qwen2.5-14B-Instruct, we achieve the best token efficiency, cutting GSM8K tokens by up to 56\% with negligible accuracy loss and, on MATH-500 (High), using a further 22\% fewer tokens than TokenSkip while improving accuracy by 7.2 points.

Overall, results on GSM8K and MATH-500 across 3B/7B/14B models show that our method consistently yields a stronger accuracy--token trade-off, establishing a new state of the art for efficient CoT reasoning.

\begin{table}[t]
\centering
\small
\setlength{\tabcolsep}{4.5pt}
\renewcommand{\arraystretch}{0.95}
\resizebox{\linewidth}{!}{
\begin{tabular}{l l c c c c}
\toprule[1.5pt]
\textbf{Dataset}  & \textbf{Setting} &
\textbf{Acc. $\uparrow$} & \textbf{Tokens $\downarrow$} &
\textbf{CR. $\downarrow$} & \textbf{TE. $\uparrow$} \\
\midrule
\multirow{4}{*}{\textbf{MATH-500}}
    & w/o HRA & 51.40 & 298.52 & 0.52 & 17.22 \\
    & w/o LPD & 52.00 & 236.98 & 0.41 & 21.94 \\
    & w/o DAG & 54.20 & 223.76 & 0.39 & 24.22 \\
    & Ours (Full) & \textbf{58.00} & 225.61 & 0.39 & \textbf{25.71} \\
\midrule
\multirow{4}{*}{\textbf{GSM8K}}
    & w/o HRA & 81.72 & 152.34 & 0.51 & 53.64 \\
    & w/o LPD & 81.35 & 137.32 & 0.46 & 59.24 \\
    & w/o DAG & 82.37 & 141.96 & 0.47 & 58.02 \\
    & Ours (Full) & \textbf{85.82} & 138.41 & 0.46 & \textbf{62.01} \\
\bottomrule[1.5pt]
\end{tabular}
}
\vspace{-0.2cm}
\caption{Ablation results on GSM8K and MATH-500. The full CtrlCoT achieves the highest accuracy and the best token efficiency (TE) on both benchmarks.}
\label{tab:distill_ablation}
\vspace{-0.15cm}
\end{table}

% \begin{table}[t]
% \centering
% \small
% \setlength{\tabcolsep}{4.5pt}
% \renewcommand{\arraystretch}{0.95}
% \resizebox{\linewidth}{!}{
% \begin{tabular}{l c l c c c c}
% \toprule[1.5pt]
% \textbf{Dataset}  & \textbf{Setting} &
% \textbf{Acc. $\uparrow$} & \textbf{Tokens $\downarrow$} &
% \textbf{CR. $\downarrow$} & \textbf{TE. $\uparrow$} \\
% \midrule
% \multirow{2}{*}{\textbf{GSM8K}}
%     & w/o D. & 69.22 & 57.74  & 0.19 & 119.89 \\
%     & w/  D. & 73.69 & 58.83  & 0.20 & \textbf{125.27} \\
% \midrule
% \multirow{2}{*}{\textbf{MATH-500}}
%    % & w/o D. & 48.00 & 198.71 & 0.35 & 24.16 \\
%    % & w/  D. & \textbf{54.00} & \textbf{187.34} & \textbf{0.33} & \textbf{28.82} \\
%    & w/o D. & 48.00 & 198.71 & 0.35 & 24.16 \\
%    & w/  D. & 54.00 & 187.34 & 0.33 & \textbf{28.82} \\
% \bottomrule[1.5pt]
% \end{tabular}
% }
% \vspace{-0.2cm}
% \caption{Impact of distillation on accuracy, token usage, and token efficiency across different budgets on GSM8K and MATH-500.}
% \label{tab:distill_ablation}
% \vspace{-0.15cm}
% \end{table}

\subsection{Ablation Study}
% Table~\ref{tab:distill_ablation} reports an ablation study on GSM8K and MATH-500. Removing any component degrades performance: excluding HRA leads to a large increase in CoT length and the lowest token efficiency, while dropping LPD or DAG reduces either accuracy or efficiency. In contrast, the full CtrlCoT achieves the best overall trade-off, attaining the highest accuracy and the highest token efficiency on both benchmarks.

% To verify the effectiveness of each component, Table~\ref{tab:distill_ablation} reports an ablation study with Qwen2.5-7B-Instruct under the High compression setting on \textsc{GSM8K} and \textsc{MATH-500}. On MATH-500, removing HRA and relying on token-level compression alone leaves substantial redundancy in the traces, resulting in longer CoTs and lower token efficiency. Removing LPD exposes the task-agnostic blindness of the pruner: without task-aware distillation, pruning is more likely to discard math-critical tokens (e.g., numbers and operators), which degrades accuracy at similar cost. Removing DAG forces the model to learn directly from fragmented pruned traces, and the resulting train--test mismatch further hurts accuracy. Overall, combining HRA, LPD, and DAG enables substantial compression while preserving correctness, and the full CtrlCoT achieves the best accuracy and token efficiency.

To verify the contribution of each component, Table~\ref{tab:distill_ablation} presents an ablation study with Qwen2.5-7B-Instruct under the High compression setting on \textsc{GSM8K} and \textsc{MATH-500}. Disabling HRA and relying mainly on token-level compression leaves more redundancy in the traces, leading to longer CoTs and lower token efficiency on both benchmarks. Without LPD, the pruner suffers from task-agnostic blindness and is more likely to drop math-critical tokens (e.g., numbers and operators), which consistently hurts accuracy at comparable cost. When DAG is absent, supervision comes directly from fragmented pruned traces, increasing the train--test mismatch and reducing accuracy.

\begin{figure}[t]
  \centering
  % (a)
  \begin{subfigure}[t]{0.48\linewidth}
    \centering
    \includegraphics[width=\linewidth]{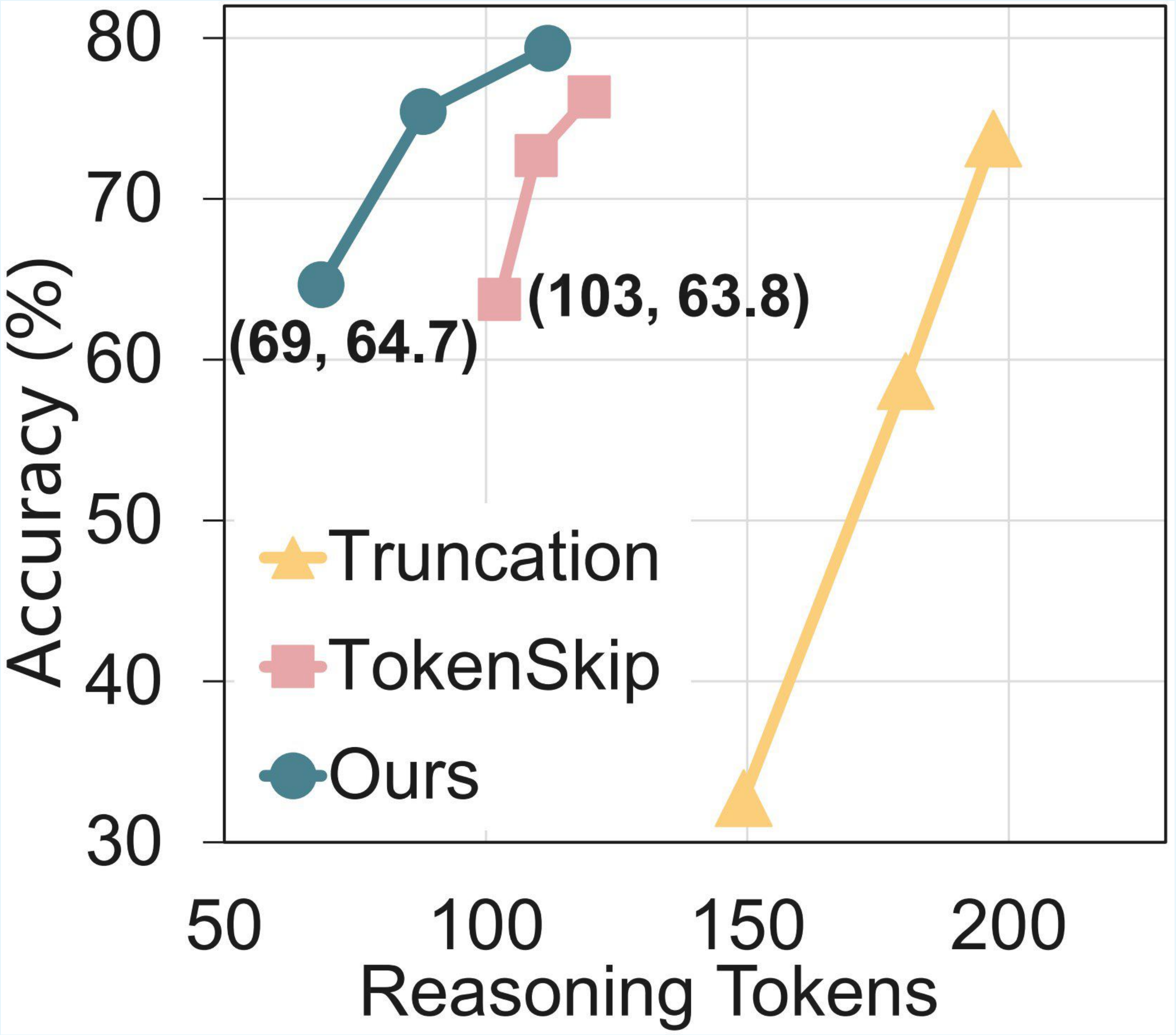}
    \caption{}\label{fig:a}
  \end{subfigure} 
  \hfill
  % (b)
  \begin{subfigure}[t]{0.48\linewidth}
    \centering
    \includegraphics[width=\linewidth]{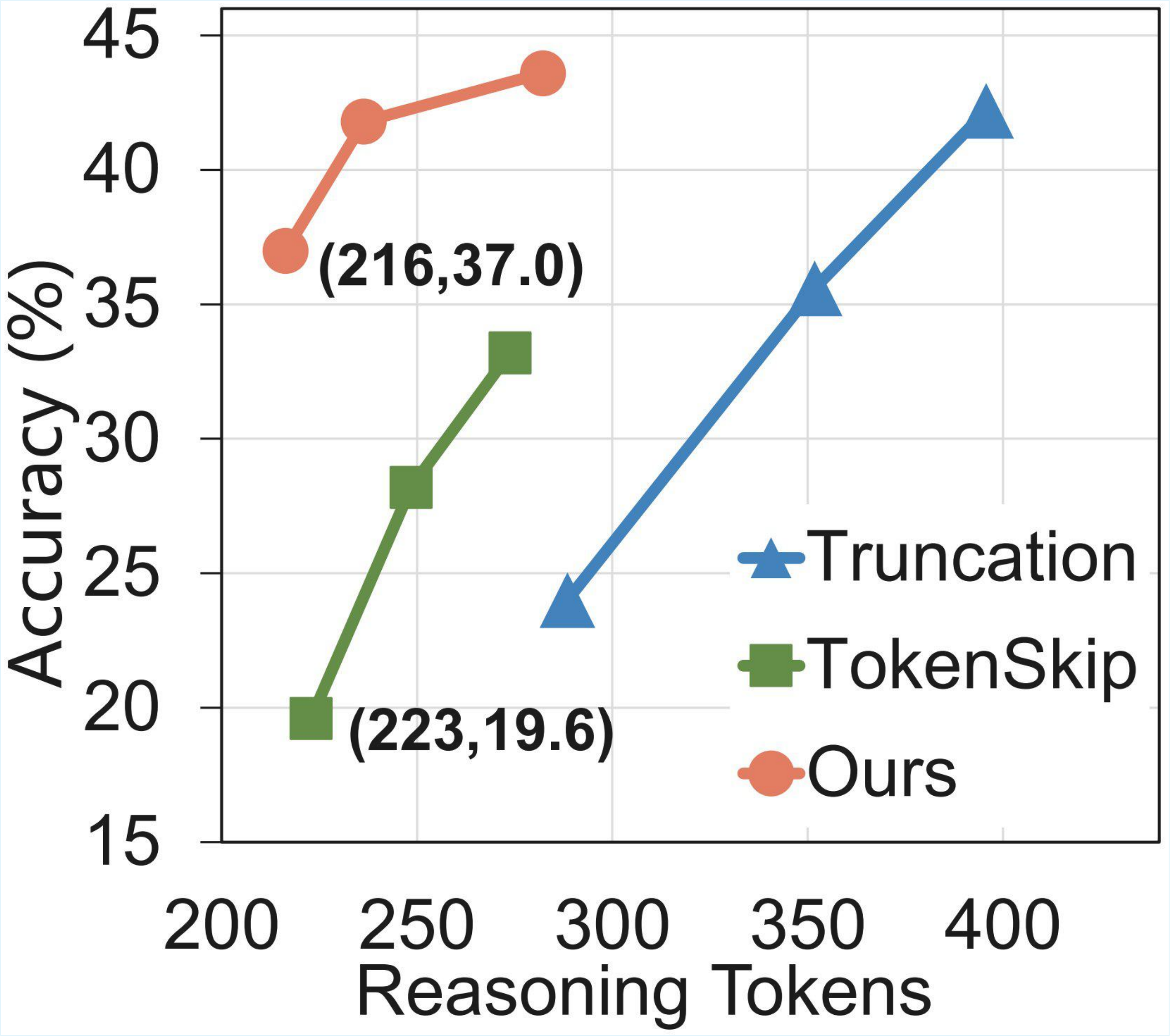}
    \caption{}\label{fig:b}
  \end{subfigure}
  \vspace{-0.2cm}
\caption{Accuracy versus CoT length on (a) GSM8K and (b) MATH-500, comparing our method with TokenSkip and Truncation under different compression strengths.}
  \label{fig:two_panels}
\vspace{-0.3cm}
\end{figure}

Overall, these results show that HRA, LPD, and DAG are complementary: together they enable substantial compression while maintaining high accuracy, and the full CtrlCoT achieves the best accuracy and token efficiency.

\subsection{In-Depth Analysis}
\noindent\textbf{Generalization to Different LLMs.} 
% \subsection{Generalization to Different LLMs.}
Figure~\ref{fig:two_panels}\subref{fig:a} and~\ref{fig:two_panels}\subref{fig:b} show that our method transfers well to LLaMA3.1-8B-Instruct, consistently achieving a better accuracy--efficiency trade-off than Truncation and TokenSkip on both GSM8K and MATH-500.
Compared to TokenSkip under Low compression strength on GSM8K, our method reduces the number of tokens by up to 33\%, while maintaining a higher accuracy.
% This results in a significant 32\% increase in token efficiency.
On MATH-500, our method increased the accuracy under a low compression strength from 19.6\% to 37.0\%, while also reducing the number of tokens.

% \noindent\textbf{Impact of Distillation.} 
% % To evaluate the effectiveness of our token pruner distillation strategy, we compare the distilled model with the original LLMLingua2 under low token budget settings on both GSM8K and MATH-500. As shown in Table~\ref{tab:distill_ablation}, the distilled model consistently achieves substantially higher accuracy while generating nearly the same number of tokens. More importantly, this accuracy gain—obtained without increasing token consumption—leads to a clear improvement in Token Efficiency. These results indicate that GPT-4–based distillation significantly enhances token pruner’s ability to identify and preserve the informative reasoning tokens within mathematical text, thereby enabling the LLM to utilize its limited token budget much more effectively.
% To evaluate our token-pruner distillation, we compare the distilled pruner with the original LLMLingua2 under low-budget settings on GSM8K and MATH-500. Table~\ref{tab:distill_ablation} shows substantially higher accuracy at nearly identical token usage, leading to improved token efficiency. This highlights the importance of a \emph{task-aware} token pruner for preserving math-critical reasoning and sustaining accuracy under tight budgets.

\begin{figure*}[t]
  \centering

  % (a) left
  \begin{subfigure}[t]{0.2835\textwidth}
    \centering
    \includegraphics[width=\linewidth]{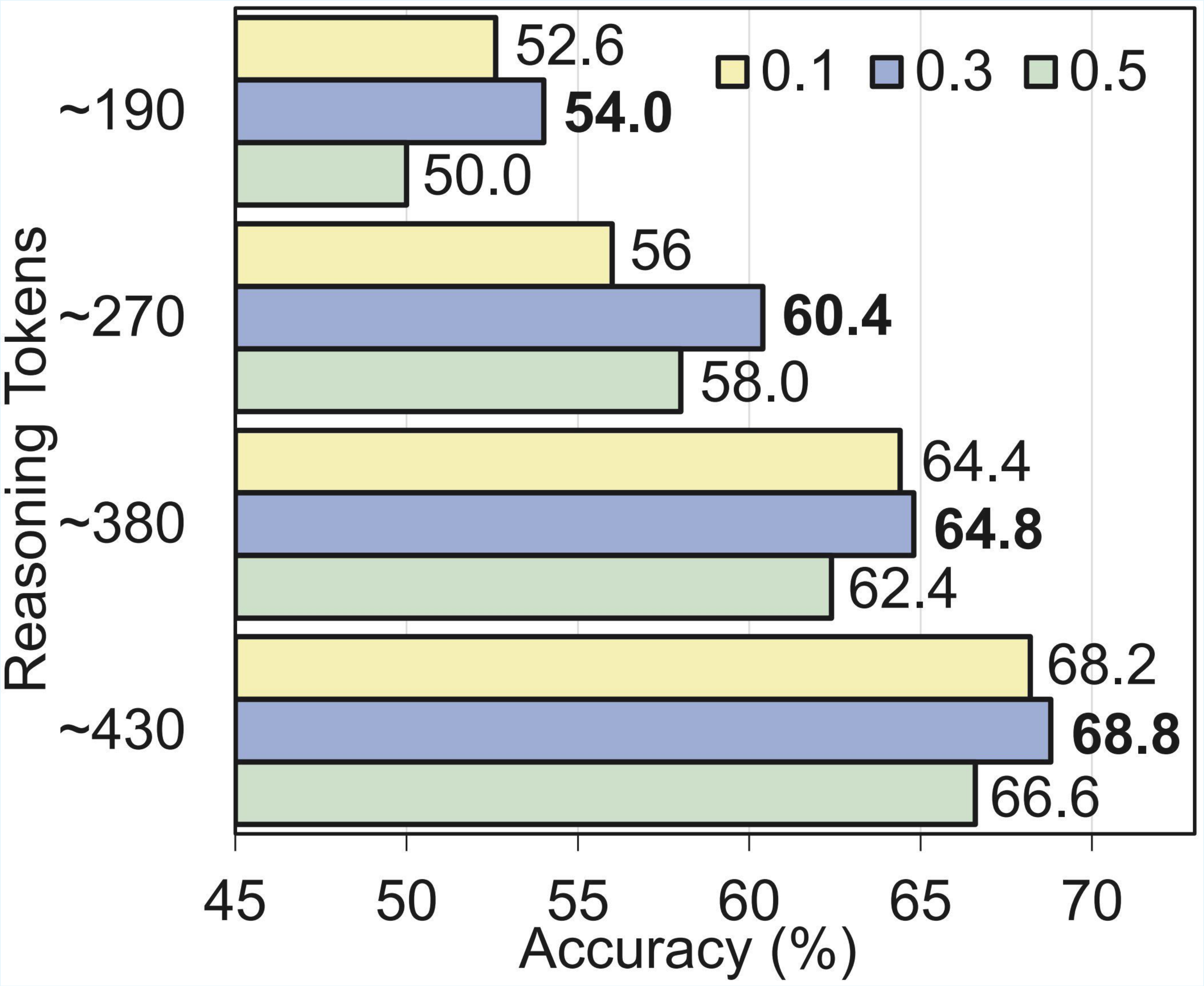}
    \caption{}\label{fig:a_}
  \end{subfigure}
  \hfill
  % (c) middle  <-- 第三个图放中间
  \begin{subfigure}[t]{0.41\textwidth}
    \centering
    \includegraphics[width=\linewidth]{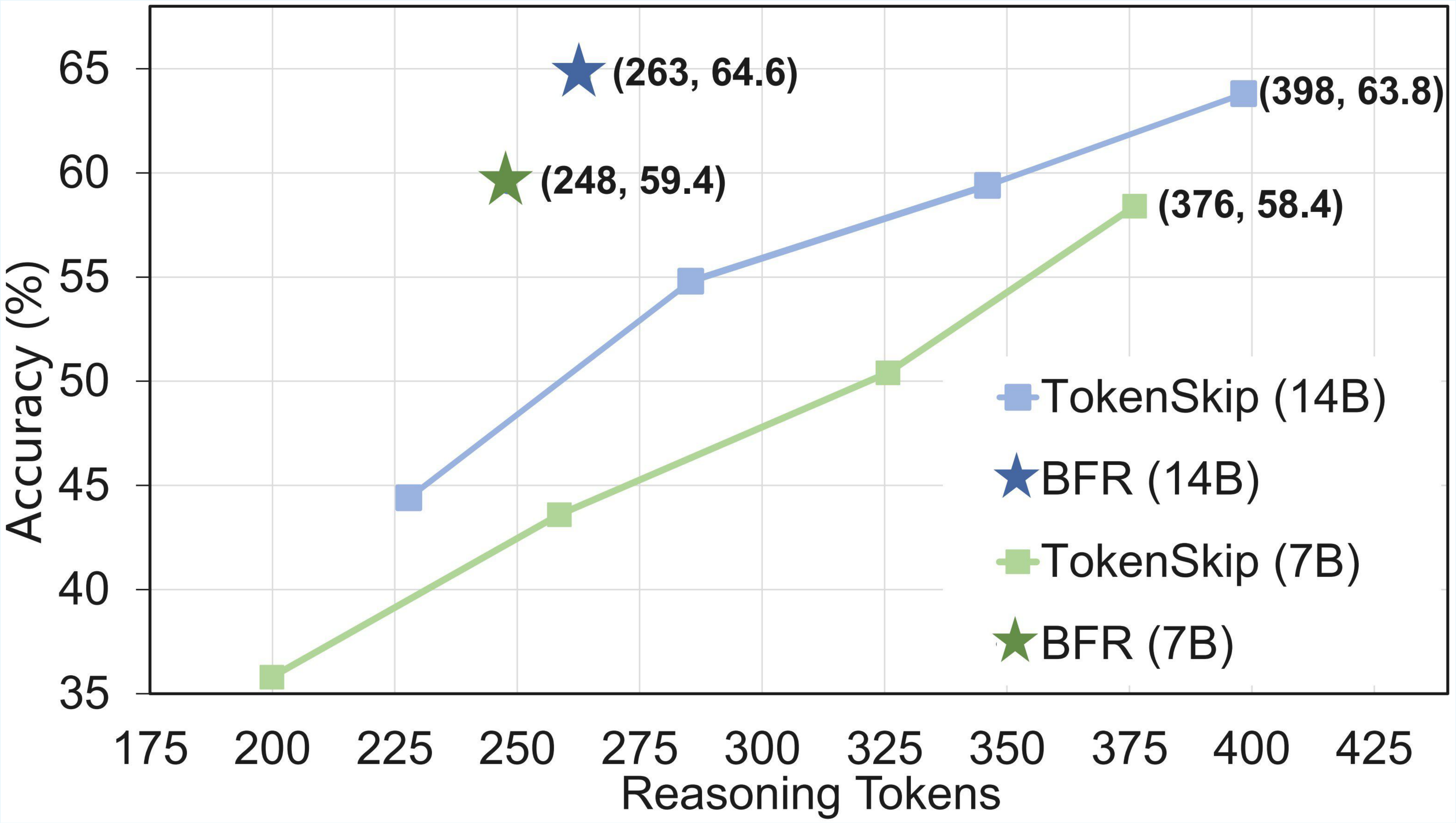}
    \caption{}\label{fig:b_}
  \end{subfigure}
  \hfill
  % (b) right
  \begin{subfigure}[t]{0.2778\textwidth}
    \centering
    \includegraphics[width=\linewidth]{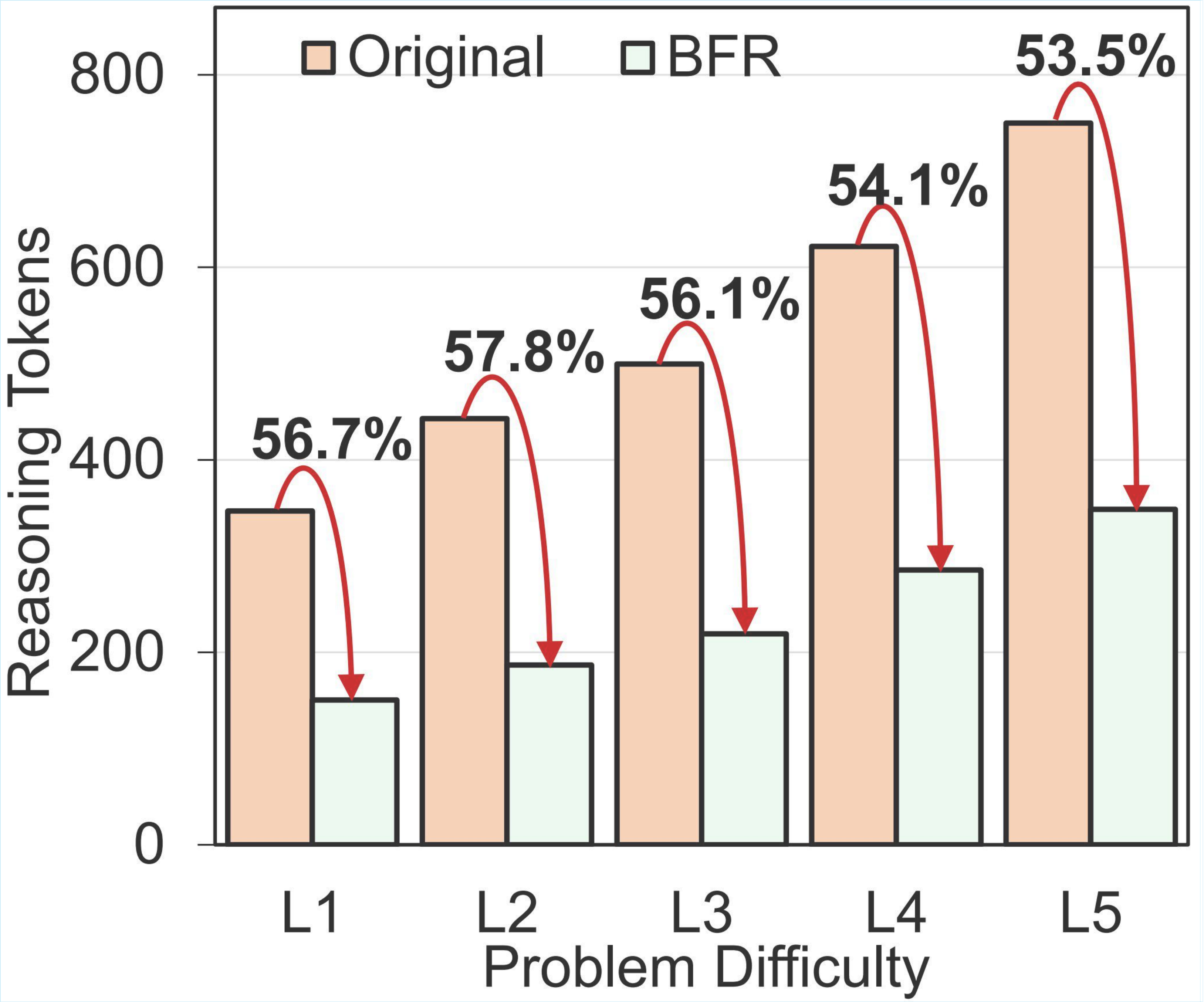}
    \caption{}\label{fig:c_}
  \end{subfigure}

  \vspace{-0.2cm}
  \caption{
  (a) The impact of the minimum compression ratio when constructing token-level compressed CoTs.
  (b) Comparison between BFR and TokenSkip on MATH-500 for Qwen2.5-7B and Qwen2.5-14B.
  (c) CoT length of the original model versus the budget-free model (BFR) on MATH-500 across difficulty levels.
  }
  \label{fig:three_in_one_row}
  \vspace{-0.3cm}
\end{figure*}

\begin{figure*}[t]
  \includegraphics[width=\linewidth]{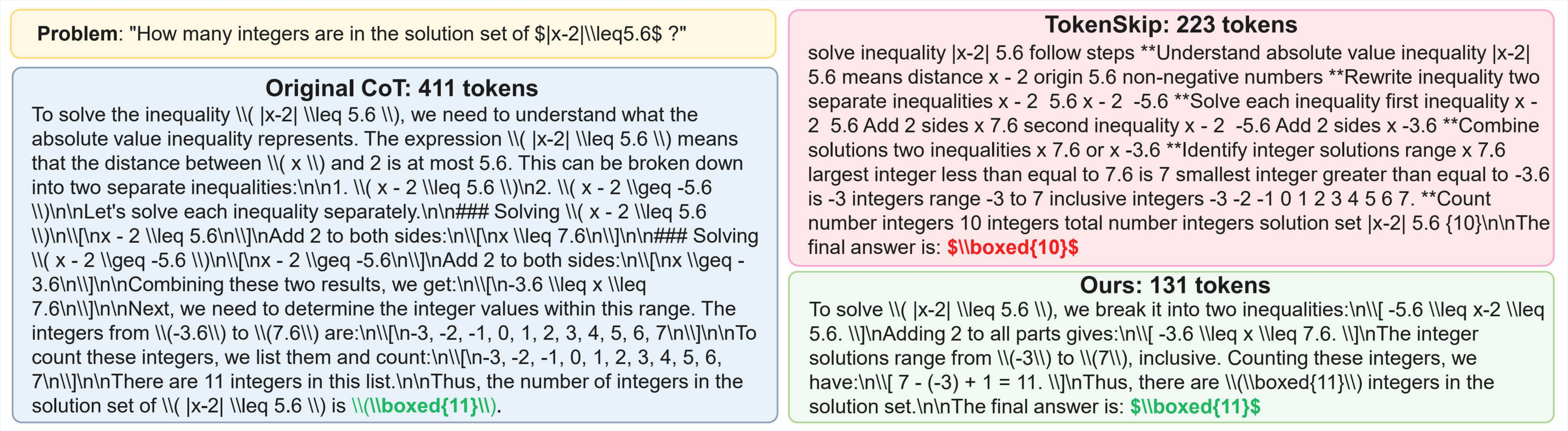}  
\caption{Case study on budgeted reasoning for an absolute-value inequality.
The original model is correct but verbose.
TokenSkip shortens the CoT but outputs an incorrect answer.
Our method is both shorter and correct.}
  
  \label{fig:case_study}
\end{figure*}

% \begin{figure}[t]
%   \centering
%   % (a)
%   \begin{subfigure}[t]{0.48\linewidth}
%     \centering
%     \includegraphics[width=\linewidth]{figs/min_ratio.pdf}
%     \caption{}\label{fig:a_}
%   \end{subfigure} 
%   \hfill
%   % (b)
%   \begin{subfigure}[t]{0.48\linewidth}
%     \centering
%     \includegraphics[width=\linewidth]{figs/optimal_diff_level.pdf}
%     \caption{}\label{fig:b_}
%   \end{subfigure}
%   \vspace{-0.2cm}
%   \caption{
% (a) The impact of the minimum compression ratio when constructing token-level compressed CoTs.
% (b) CoT length of the original model versus the budget-free model (BFR) on MATH-500 across difficulty levels. 
% }
%   \label{fig:two_panels}
% \vspace{-0.3cm}
% \end{figure}

\paragraph{Effect of Minimum Compression Ratio.}
% 相同预算下的acc
% In both training and inference of MCG, we set the minimum compression ratio to 0.3, and Figure~\ref{fig:a_} studies its impact. A minimum ratio of 0.3 yields the best performance across CoT budgets, especially for short CoTs. This may be because if the minimum ratio is too low, the CoTs may lose key information, making them hard to understand and undermining the model’s reasoning ability. If the minimum ratio is too high, the CoTs remain overly redundant, and the model is less likely to learn to generate concise reasoning. Overall, 0.3 strikes a better balance and delivers the strongest accuracy--token trade-off.

For both training and inference of MCG, the minimum compression ratio is set to 0.3, and Figure~\ref{fig:a_} analyzes its effect. A minimum ratio of 0.3 performs best across budgets, particularly for short CoTs. When the ratio is lower, pruning tends to remove essential information and hurts reasoning; when it is higher, the CoTs remain redundant and provide weaker supervision for concise generation. Overall, 0.3 delivers the best overall balance.

\paragraph{Budget-Free CoT Generation.}

We compare our budget-free model (BFR) with TokenSkip on MATH-500 using Qwen2.5-7B-Instruct and Qwen2.5-14B-Instruct. As shown in Figure~\ref{fig:b_}, BFR achieves higher accuracy while using roughly 130 fewer reasoning tokens than TokenSkip on both backbones, placing it clearly above the TokenSkip curves and indicating substantially higher token efficiency.

\noindent\textbf{CoT Length Analysis of BFR.}
Figure~\ref{fig:c_} reports CoT lengths of the original Qwen2.5-7B-Instruct model and BFR on MATH-500 across difficulty levels. BFR consistently generates shorter CoTs than the original model at every level, and the relative reduction is larger on easier problems. This may be because easier questions require fewer essential steps, allowing more aggressive compression, whereas harder questions demand more tokens to support complex multi-step reasoning.

% \begin{table}[t]
% \centering
% \small
% \setlength{\tabcolsep}{4pt}
% \begin{tabular}{l c l c c c c}
% \toprule
% \textbf{Dataset} & \textbf{Budget} & \textbf{Setting} &
% \textbf{Acc. $\uparrow$} & \textbf{Tokens $\downarrow$} &
% \textbf{CR. $\downarrow$} & \textbf{TE. $\uparrow$} \\
% \midrule

% \multirow{4}{*}{\textbf{MATH-500}}
%   & 100 & w/o Distill & 48.00 & 198.71 & 0.35 & 24.16 \\
%   & 100 & w/  Distill & \textbf{54.00} & \textbf{187.34} & \textbf{0.33} & \textbf{28.82} \\
%   & 200 & w/o Distill & 58.00 & \textbf{263.40} & \textbf{0.46} & \textbf{22.02} \\
%   & 200 & w/  Distill & \textbf{60.40} & 267.07 & \textbf{0.46} & \textbf{22.62} \\
% \midrule

% \multirow{4}{*}{\textbf{GSM8K}}
%   & 50  & w/o Distill & 69.22 & \textbf{57.74}  & \textbf{0.19} & 119.89 \\
%   & 50  & w/  Distill & \textbf{73.69} & 58.83  & 0.20 & \textbf{125.27} \\
%   & 100 & w/o Distill & 81.73 & \textbf{109.47} & \textbf{0.37} & 74.66 \\
%   & 100 & w/  Distill & \textbf{83.17} & 109.80 & \textbf{0.37} & \textbf{75.74} \\
% \bottomrule
% \end{tabular}
% \caption{Impact of distillation on accuracy, token usage, and token efficiency across different budgets on GSM8K and MATH-500.}
% \label{tab:distill_ablation}
% \end{table}

\paragraph{Case Study.}
% Figure~\ref{fig:case_study} shows an absolute-value inequality example.
% The original model gives the correct answer but produces a verbose CoT with 411 tokens.
% TokenSkip shortens the CoT to 223 tokens, yet makes a counting mistake and outputs an incorrect result.
% In contrast, our method generates a much shorter CoT with 131 tokens while preserving the essential steps and staying correct.
% This example demonstrates that token-level compression alone can degrade accuracy, whereas our approach maintains faithfulness under stronger compression.

Figure~\ref{fig:case_study} presents an absolute-value inequality example. The original model is correct but verbose (411 tokens). TokenSkip is shorter (223) but makes a counting error. In contrast, our method remains correct with a much shorter rationale (131), illustrating that token-only compression can hurt accuracy while ours preserves reliability under stronger compression.

\section{Conclusion}
This paper presents CtrlCoT, a dual-granularity CoT compression framework that coordinates semantic abstraction and token-level pruning to reduce reasoning cost without sacrificing correctness. CtrlCoT addresses three key challenges in joint compression---sequential dependency, task-agnostic pruning, and distribution mismatch---via Hierarchical Reasoning Abstraction, Logic-Preserving Distillation, and Distribution-Alignment Generation. Experiments on GSM8K and MATH-500 across multiple model scales show that CtrlCoT consistently achieves higher accuracy at comparable or shorter CoT lengths than strong baselines, establishing a new state of the art for efficient reasoning. Future work includes extending CtrlCoT beyond mathematical reasoning to broader domains and exploring finer-grained compression strategies.

\section*{Limitations}
Our method depends on the backbone model to follow budget instructions during generation. However, due to limited controllability of current LLMs, the produced CoT length can still deviate from the token budget specified in the prompt. This mismatch is more visible on weaker backbones. Closing this budget--length mismatch remains an important direction for future work.

\section*{Ethics Statement}
All datasets in our experiments are publicly released and were annotated through human interactions conducted in English. We took measures to protect user privacy during the annotation process, and the data contain no personal or identifying information. The scientific artifacts used in this work are provided for research under permissive licenses, and we use them in accordance with their intended scope. Accordingly, we believe this work satisfies ACL ethical requirements.

% \section*{Acknowledgments}

% This document has been adapted
% by Steven Bethard, Ryan Cotterell and Rui Yan
% from the instructions for earlier ACL and NAACL proceedings, including those for
% ACL 2019 by Douwe Kiela and Ivan Vuli\'{c},
% NAACL 2019 by Stephanie Lukin and Alla Roskovskaya,
% ACL 2018 by Shay Cohen, Kevin Gimpel, and Wei Lu,
% NAACL 2018 by Margaret Mitchell and Stephanie Lukin,
% Bib\TeX{} suggestions for (NA)ACL 2017/2018 from Jason Eisner,
% ACL 2017 by Dan Gildea and Min-Yen Kan,
% NAACL 2017 by Margaret Mitchell,
% ACL 2012 by Maggie Li and Michael White,
% ACL 2010 by Jing-Shin Chang and Philipp Koehn,
% ACL 2008 by Johanna D. Moore, Simone Teufel, James Allan, and Sadaoki Furui,
% ACL 2005 by Hwee Tou Ng and Kemal Oflazer,
% ACL 2002 by Eugene Charniak and Dekang Lin,
% and earlier ACL and EACL formats written by several people, including
% John Chen, Henry S. Thompson and Donald Walker.
% Additional elements were taken from the formatting instructions of the \emph{International Joint Conference on Artificial Intelligence} and the \emph{Conference on Computer Vision and Pattern Recognition}.

% Bibliography entries for the entire Anthology, followed by custom entries
%\bibliography{custom,anthology-overleaf-1,anthology-overleaf-2}

% Custom bibliography entries only
% \bibliography{custom}
\bibliography{main}

@misc{xia2025tokenskip,
      title={TokenSkip: Controllable Chain-of-Thought Compression in LLMs}, 
      author={Heming Xia and Yongqi Li and Chak Tou Leong and Wenjie Wang and Wenjie Li},
      year={2025},
      eprint={2502.12067},
      archivePrefix={arXiv},
      primaryClass={cs.CL},
      url={https://arxiv.org/abs/2502.12067}, 
}

@inproceedings{NEURIPS2022_9d560961,
 author = {Wei, Jason and Wang, Xuezhi and Schuurmans, Dale and Bosma, Maarten and ichter, brian and Xia, Fei and Chi, Ed and Le, Quoc V and Zhou, Denny},
 booktitle = {Advances in Neural Information Processing Systems},
 editor = {S. Koyejo and S. Mohamed and A. Agarwal and D. Belgrave and K. Cho and A. Oh},
 pages = {24824--24837},
 publisher = {Curran Associates, Inc.},
 title = {Chain-of-Thought Prompting Elicits Reasoning in Large Language Models},
 url = {https://proceedings.neurips.cc/paper_files/paper/2022/file/9d5609613524ecf4f15af0f7b31abca4-Paper-Conference.pdf},
 volume = {35},
 year = {2022}
}

@misc{muennighoff2025s1simpletesttimescaling,
      title={s1: Simple test-time scaling}, 
      author={Niklas Muennighoff and Zitong Yang and Weijia Shi and Xiang Lisa Li and Li Fei-Fei and Hannaneh Hajishirzi and Luke Zettlemoyer and Percy Liang and Emmanuel Candès and Tatsunori Hashimoto},
      year={2025},
      eprint={2501.19393},
      archivePrefix={arXiv},
      primaryClass={cs.CL},
      url={https://arxiv.org/abs/2501.19393}, 
}

@misc{ma2025cotvalvelengthcompressiblechainofthoughttuning,
      title={CoT-Valve: Length-Compressible Chain-of-Thought Tuning}, 
      author={Xinyin Ma and Guangnian Wan and Runpeng Yu and Gongfan Fang and Xinchao Wang},
      year={2025},
      eprint={2502.09601},
      archivePrefix={arXiv},
      primaryClass={cs.AI},
      url={https://arxiv.org/abs/2502.09601}, 
}

@inproceedings{pan2024llmlingua,
  title={LLMLingua-2: Data Distillation for Efficient and Faithful Task-Agnostic Prompt Compression},
  author={Pan, Zhuoshi and Wu, Qianhui and Jiang, Huiqiang and Xia, Menglin and Luo, Xufang and Zhang, Jue and Lin, Qingwei and R{\"u}hle, Victor and Yang, Yuqing and Lin, Chin-Yew and others},
  booktitle={ACL (Findings)},
  year={2024}
}

@misc{yang2024qwen2technicalreport,
      title={Qwen2 Technical Report}, 
      author={An Yang and Baosong Yang and Binyuan Hui and Bo Zheng and Bowen Yu and Chang Zhou and Chengpeng Li and Chengyuan Li and Dayiheng Liu and Fei Huang and Guanting Dong and Haoran Wei and Huan Lin and Jialong Tang and Jialin Wang and Jian Yang and Jianhong Tu and Jianwei Zhang and Jianxin Ma and Jianxin Yang and Jin Xu and Jingren Zhou and Jinze Bai and Jinzheng He and Junyang Lin and Kai Dang and Keming Lu and Keqin Chen and Kexin Yang and Mei Li and Mingfeng Xue and Na Ni and Pei Zhang and Peng Wang and Ru Peng and Rui Men and Ruize Gao and Runji Lin and Shijie Wang and Shuai Bai and Sinan Tan and Tianhang Zhu and Tianhao Li and Tianyu Liu and Wenbin Ge and Xiaodong Deng and Xiaohuan Zhou and Xingzhang Ren and Xinyu Zhang and Xipin Wei and Xuancheng Ren and Xuejing Liu and Yang Fan and Yang Yao and Yichang Zhang and Yu Wan and Yunfei Chu and Yuqiong Liu and Zeyu Cui and Zhenru Zhang and Zhifang Guo and Zhihao Fan},
      year={2024},
      eprint={2407.10671},
      archivePrefix={arXiv},
      primaryClass={cs.CL},
      url={https://arxiv.org/abs/2407.10671}, 
}

@misc{cobbe2021trainingverifierssolvemath,
      title={Training Verifiers to Solve Math Word Problems}, 
      author={Karl Cobbe and Vineet Kosaraju and Mohammad Bavarian and Mark Chen and Heewoo Jun and Lukasz Kaiser and Matthias Plappert and Jerry Tworek and Jacob Hilton and Reiichiro Nakano and Christopher Hesse and John Schulman},
      year={2021},
      eprint={2110.14168},
      archivePrefix={arXiv},
      primaryClass={cs.LG},
      url={https://arxiv.org/abs/2110.14168}, 
}

@inproceedings{DBLP:conf/nips/HendrycksBKABTS21,
  author       = {Dan Hendrycks and
                  Collin Burns and
                  Saurav Kadavath and
                  Akul Arora and
                  Steven Basart and
                  Eric Tang and
                  Dawn Song and
                  Jacob Steinhardt},
  editor       = {Joaquin Vanschoren and
                  Sai{-}Kit Yeung},
  title        = {Measuring Mathematical Problem Solving With the {MATH} Dataset},
  booktitle    = {Proceedings of the Neural Information Processing Systems Track on
                  Datasets and Benchmarks 1, NeurIPS Datasets and Benchmarks 2021, December
                  2021, virtual},
  year         = {2021},
  url          = {https://datasets-benchmarks-proceedings.neurips.cc/paper/2021/hash/be83ab3ecd0db773eb2dc1b0a17836a1-Abstract-round2.html},
  bibsource    = {dblp computer science bibliography, https://dblp.org}
}

@inproceedings{DBLP:conf/iclr/LightmanKBEBLLS24,
  author       = {Hunter Lightman and
                  Vineet Kosaraju and
                  Yuri Burda and
                  Harrison Edwards and
                  Bowen Baker and
                  Teddy Lee and
                  Jan Leike and
                  John Schulman and
                  Ilya Sutskever and
                  Karl Cobbe},
  title        = {Let's Verify Step by Step},
  booktitle    = {The Twelfth International Conference on Learning Representations,
                  {ICLR} 2024, Vienna, Austria, May 7-11, 2024},
  publisher    = {OpenReview.net},
  year         = {2024},
  url          = {https://openreview.net/forum?id=v8L0pN6EOi},
  bibsource    = {dblp computer science bibliography, https://dblp.org}
}

@inproceedings{jiang2023llmlingua,
  title={LLMLingua: Compressing Prompts for Accelerated Inference of Large Language Models},
  author={Jiang, Huiqiang and Wu, Qianhui and Lin, Chin-Yew and Yang, Yuqing and Qiu, Lili},
  booktitle={Proceedings of the 2023 Conference on Empirical Methods in Natural Language Processing},
  pages={13358--13376},
  year={2023}
}

@misc{sui2025stopoverthinkingsurveyefficient,
      title={Stop Overthinking: A Survey on Efficient Reasoning for Large Language Models}, 
      author={Yang Sui and Yu-Neng Chuang and Guanchu Wang and Jiamu Zhang and Tianyi Zhang and Jiayi Yuan and Hongyi Liu and Andrew Wen and Shaochen Zhong and Hanjie Chen and Xia Hu},
      year={2025},
      eprint={2503.16419},
      archivePrefix={arXiv},
      primaryClass={cs.CL},
      url={https://arxiv.org/abs/2503.16419}, 
}

@inproceedings{DBLP:conf/aaai/KangSCZ25,
  author       = {Yu Kang and
                  Xianghui Sun and
                  Liangyu Chen and
                  Wei Zou},
  editor       = {Toby Walsh and
                  Julie Shah and
                  Zico Kolter},
  title        = {C3oT: Generating Shorter Chain-of-Thought Without Compromising Effectiveness},
  booktitle    = {AAAI-25, Sponsored by the Association for the Advancement of Artificial
                  Intelligence, February 25 - March 4, 2025, Philadelphia, PA, {USA}},
  pages        = {24312--24320},
  publisher    = {{AAAI} Press},
  year         = {2025},
  url          = {https://doi.org/10.1609/aaai.v39i23.34608},
  doi          = {10.1609/AAAI.V39I23.34608},
  bibsource    = {dblp computer science bibliography, https://dblp.org}
}

@article{hu2022lora,
  title={Lora: Low-rank adaptation of large language models.},
  author={Hu, Edward J and Shen, Yelong and Wallis, Phillip and Allen-Zhu, Zeyuan and Li, Yuanzhi and Wang, Shean and Wang, Lu and Chen, Weizhu and others},
  journal={ICLR},
  volume={1},
  number={2},
  pages={3},
  year={2022}
}

@inproceedings{zheng2024llamafactory,
  title={LlamaFactory: Unified Efficient Fine-Tuning of 100+ Language Models},
  author={Yaowei Zheng and Richong Zhang and Junhao Zhang and Yanhan Ye and Zheyan Luo and Zhangchi Feng and Yongqiang Ma},
  booktitle={Proceedings of the 62nd Annual Meeting of the Association for Computational Linguistics (Volume 3: System Demonstrations)},
  address={Bangkok, Thailand},
  publisher={Association for Computational Linguistics},
  year={2024},
  url={http://arxiv.org/abs/2403.13372}
}

@misc{deepseek-math,
      title={DeepSeekMath: Pushing the Limits of Mathematical Reasoning in Open Language Models}, 
      author={Zhihong Shao and Peiyi Wang and Qihao Zhu and Runxin Xu and Junxiao Song and Xiao Bi and Haowei Zhang and Mingchuan Zhang and Y. K. Li and Y. Wu and Daya Guo},
      year={2024},
      eprint={2402.03300},
      archivePrefix={arXiv},
      primaryClass={cs.CL},
      url={https://arxiv.org/abs/2402.03300}, 
}

@inproceedings{
hao2025training,
title={Training Large Language Models to Reason in a Continuous Latent Space},
author={Shibo Hao and Sainbayar Sukhbaatar and DiJia Su and Xian Li and Zhiting Hu and Jason E Weston and Yuandong Tian},
booktitle={Second Conference on Language Modeling},
year={2025},
url={https://openreview.net/forum?id=Itxz7S4Ip3}
}

@inproceedings{shen-etal-2025-codi,
    title = "{CODI}: Compressing Chain-of-Thought into Continuous Space via Self-Distillation",
    author = "Shen, Zhenyi  and
      Yan, Hanqi  and
      Zhang, Linhai  and
      Hu, Zhanghao  and
      Du, Yali  and
      He, Yulan",
    editor = "Christodoulopoulos, Christos  and
      Chakraborty, Tanmoy  and
      Rose, Carolyn  and
      Peng, Violet",
    booktitle = "Proceedings of the 2025 Conference on Empirical Methods in Natural Language Processing",
    month = nov,
    year = "2025",
    address = "Suzhou, China",
    publisher = "Association for Computational Linguistics",
    url = "https://aclanthology.org/2025.emnlp-main.36/",
    doi = "10.18653/v1/2025.emnlp-main.36",
    pages = "677--693",
    ISBN = "979-8-89176-332-6"
}

@article{cheng2024compressed,
  title={Compressed chain of thought: Efficient reasoning through dense representations},
  author={Cheng, Jeffrey and Van Durme, Benjamin},
  journal={arXiv preprint arXiv:2412.13171},
  year={2024}
}

@article{team2025kimi,
  title={Kimi k1. 5: Scaling reinforcement learning with llms},
  author={Team, Kimi and Du, Angang and Gao, Bofei and Xing, Bowei and Jiang, Changjiu and Chen, Cheng and Li, Cheng and Xiao, Chenjun and Du, Chenzhuang and Liao, Chonghua and others},
  journal={arXiv preprint arXiv:2501.12599},
  year={2025}
}

@article{luo2025o1,
  title={O1-pruner: Length-harmonizing fine-tuning for o1-like reasoning pruning},
  author={Luo, Haotian and Shen, Li and He, Haiying and Wang, Yibo and Liu, Shiwei and Li, Wei and Tan, Naiqiang and Cao, Xiaochun and Tao, Dacheng},
  journal={arXiv preprint arXiv:2501.12570},
  year={2025}
}

@article{aggarwal2025l1,
  title={L1: Controlling how long a reasoning model thinks with reinforcement learning},
  author={Aggarwal, Pranjal and Welleck, Sean},
  journal={arXiv preprint arXiv:2503.04697},
  year={2025}
}

@article{sun2024fast,
  title={Fast best-of-n decoding via speculative rejection},
  author={Sun, Hanshi and Haider, Momin and Zhang, Ruiqi and Yang, Huitao and Qiu, Jiahao and Yin, Ming and Wang, Mengdi and Bartlett, Peter and Zanette, Andrea},
  journal={Advances in Neural Information Processing Systems},
  volume={37},
  pages={32630--32652},
  year={2024}
}

@inproceedings{ding-etal-2025-dynamic,
    title = "Dynamic Parallel Tree Search for Efficient {LLM} Reasoning",
    author = "Ding, Yifu  and
      Jiang, Wentao  and
      Liu, Shunyu  and
      Jing, Yongcheng  and
      Guo, Jinyang  and
      Wang, Yingjie  and
      Zhang, Jing  and
      Wang, Zengmao  and
      Liu, Ziwei  and
      Du, Bo  and
      Liu, Xianglong  and
      Tao, Dacheng",
    editor = "Che, Wanxiang  and
      Nabende, Joyce  and
      Shutova, Ekaterina  and
      Pilehvar, Mohammad Taher",
    booktitle = "Proceedings of the 63rd Annual Meeting of the Association for Computational Linguistics (Volume 1: Long Papers)",
    month = jul,
    year = "2025",
    address = "Vienna, Austria",
    publisher = "Association for Computational Linguistics",
    url = "https://aclanthology.org/2025.acl-long.550/",
    doi = "10.18653/v1/2025.acl-long.550",
    pages = "11233--11252",
    ISBN = "979-8-89176-251-0"
}

@article{wang2025sampling,
  title={Sampling-efficient test-time scaling: Self-estimating the best-of-n sampling in early decoding},
  author={Wang, Yiming and Zhang, Pei and Huang, Siyuan and Yang, Baosong and Zhang, Zhuosheng and Huang, Fei and Wang, Rui},
  journal={arXiv preprint arXiv:2503.01422},
  year={2025}
}

@article{
sui2025stop,
title={Stop Overthinking: A Survey on Efficient Reasoning for Large Language Models},
author={Yang Sui and Yu-Neng Chuang and Guanchu Wang and Jiamu Zhang and Tianyi Zhang and Jiayi Yuan and Hongyi Liu and Andrew Wen and Shaochen Zhong and Na Zou and Hanjie Chen and Xia Hu},
journal={Transactions on Machine Learning Research},
issn={2835-8856},
year={2025},
url={https://openreview.net/forum?id=HvoG8SxggZ},
note={}
}

@inproceedings{MLSYS2024_42a452cb,
 author = {Lin, Ji and Tang, Jiaming and Tang, Haotian and Yang, Shang and Chen, Wei-Ming and Wang, Wei-Chen and Xiao, Guangxuan and Dang, Xingyu and Gan, Chuang and Han, Song},
 booktitle = {Proceedings of Machine Learning and Systems},
 editor = {P. Gibbons and G. Pekhimenko and C. De Sa},
 pages = {87--100},
 title = {AWQ: Activation-aware Weight Quantization for On-Device LLM Compression and Acceleration},
 url = {https://proceedings.mlsys.org/paper_files/paper/2024/file/42a452cbafa9dd64e9ba4aa95cc1ef21-Paper-Conference.pdf},
 volume = {6},
 year = {2024}
}

@misc{dao2023flashattention2fasterattentionbetter,
      title={FlashAttention-2: Faster Attention with Better Parallelism and Work Partitioning}, 
      author={Tri Dao},
      year={2023},
      eprint={2307.08691},
      archivePrefix={arXiv},
      primaryClass={cs.LG},
      url={https://arxiv.org/abs/2307.08691}, 
}

@inproceedings{ainslie-etal-2023-gqa,
    title = "{GQA}: Training Generalized Multi-Query Transformer Models from Multi-Head Checkpoints",
    author = "Ainslie, Joshua  and
      Lee-Thorp, James  and
      de Jong, Michiel  and
      Zemlyanskiy, Yury  and
      Lebron, Federico  and
      Sanghai, Sumit",
    editor = "Bouamor, Houda  and
      Pino, Juan  and
      Bali, Kalika",
    booktitle = "Proceedings of the 2023 Conference on Empirical Methods in Natural Language Processing",
    month = dec,
    year = "2023",
    address = "Singapore",
    publisher = "Association for Computational Linguistics",
    url = "https://aclanthology.org/2023.emnlp-main.298/",
    doi = "10.18653/v1/2023.emnlp-main.298",
    pages = "4895--4901"
}

@inproceedings{
chen2025do,
title={Do {NOT} Think That Much for 2+3=? On the Overthinking of Long Reasoning Models},
author={Xingyu Chen and Jiahao Xu and Tian Liang and Zhiwei He and Jianhui Pang and Dian Yu and Linfeng Song and Qiuzhi Liu and Mengfei Zhou and Zhuosheng Zhang and Rui Wang and Zhaopeng Tu and Haitao Mi and Dong Yu},
booktitle={Forty-second International Conference on Machine Learning},
year={2025},
url={https://openreview.net/forum?id=MSbU3L7V00}
}

@inproceedings{
fan2025missing,
title={Missing Premise exacerbates Overthinking: Are Reasoning Models losing Critical Thinking Skill?},
author={Chenrui Fan and Ming Li and Lichao Sun and Tianyi Zhou},
booktitle={Second Conference on Language Modeling},
year={2025},
url={https://openreview.net/forum?id=ufozo2Wc9e}
}

@article{qu2025survey,
  title={A Survey of Efficient Reasoning for Large Reasoning Models: Language, Multimodality, and Beyond},
  author={Qu, Xiaoye and Li, Yafu and Su, Zhaochen and Sun, Weigao and Yan, Jianhao and Liu, Dongrui and Cui, Ganqu and Liu, Daizong and Liang, Shuxian and He, Junxian and others},
  journal={arXiv preprint arXiv:2503.21614},
  year={2025}
}

@inproceedings{DBLP:conf/iclr/LuoSX0LTGLCT025,
  author       = {Haipeng Luo and
                  Qingfeng Sun and
                  Can Xu and
                  Pu Zhao and
                  Jian{-}Guang Lou and
                  Chongyang Tao and
                  Xiubo Geng and
                  Qingwei Lin and
                  Shifeng Chen and
                  Yansong Tang and
                  Dongmei Zhang},
  title        = {WizardMath: Empowering Mathematical Reasoning for Large Language Models
                  via Reinforced Evol-Instruct},
  booktitle    = {The Thirteenth International Conference on Learning Representations,
                  {ICLR} 2025, Singapore, April 24-28, 2025},
  publisher    = {OpenReview.net},
  year         = {2025},
  url          = {https://openreview.net/forum?id=mMPMHWOdOy},
  bibsource    = {dblp computer science bibliography, https://dblp.org}
}

@inproceedings{
wang2025logictree,
title={LogicTree: Improving Complex Reasoning of {LLM}s via Instantiated Multi-step Synthetic Logical Data},
author={Zehao Wang and Lin Yang and Jie Wang and Kehan Wang and Hanzhu Chen and Bin Wang and Jianye HAO and Defu Lian and Bin Li and Enhong Chen},
booktitle={The Thirty-ninth Annual Conference on Neural Information Processing Systems},
year={2025},
url={https://openreview.net/forum?id=z4AMrCOetn}
}

@inproceedings{leviathan2023fast,
  title={Fast inference from transformers via speculative decoding},
  author={Leviathan, Yaniv and Kalman, Matan and Matias, Yossi},
  booktitle={International Conference on Machine Learning},
  pages={19274--19286},
  year={2023},
  organization={PMLR}
}

@article{liu2024minicache,
  title={Minicache: Kv cache compression in depth dimension for large language models},
  author={Liu, Akide and Liu, Jing and Pan, Zizheng and He, Yefei and Haffari, Gholamreza and Zhuang, Bohan},
  journal={Advances in Neural Information Processing Systems},
  volume={37},
  pages={139997--140031},
  year={2024}
}

@article{lin2024duquant,
  title={Duquant: Distributing outliers via dual transformation makes stronger quantized llms},
  author={Lin, Haokun and Xu, Haobo and Wu, Yichen and Cui, Jingzhi and Zhang, Yingtao and Mou, Linzhan and Song, Linqi and Sun, Zhenan and Wei, Ying},
  journal={Advances in Neural Information Processing Systems},
  volume={37},
  pages={87766--87800},
  year={2024}
}

@article{alomrani2025reasoning,
  title={Reasoning on a Budget: A Survey of Adaptive and Controllable Test-Time Compute in LLMs},
  author={Alomrani, Mohammad Ali and Zhang, Yingxue and Li, Derek and Sun, Qianyi and Pal, Soumyasundar and Zhang, Zhanguang and Hu, Yaochen and Ajwani, Rohan Deepak and Valkanas, Antonios and Karimi, Raika and others},
  journal={arXiv preprint arXiv:2507.02076},
  year={2025}
}

@misc{grattafiori2024llama3herdmodels,
      title={The Llama 3 Herd of Models}, 
      author={Aaron Grattafiori and Abhimanyu Dubey and Abhinav Jauhri and Abhinav Pandey and Abhishek Kadian and Ahmad Al-Dahle and Aiesha Letman and Akhil Mathur and Alan Schelten and Alex Vaughan and Amy Yang and Angela Fan and Anirudh Goyal and Anthony Hartshorn and Aobo Yang and Archi Mitra and Archie Sravankumar and Artem Korenev and Arthur Hinsvark and Arun Rao and Aston Zhang and Aurelien Rodriguez and Austen Gregerson and Ava Spataru and Baptiste Roziere and Bethany Biron and Binh Tang and Bobbie Chern and Charlotte Caucheteux and Chaya Nayak and Chloe Bi and Chris Marra and Chris McConnell and Christian Keller and Christophe Touret and Chunyang Wu and Corinne Wong and Cristian Canton Ferrer and Cyrus Nikolaidis and Damien Allonsius and Daniel Song and Danielle Pintz and Danny Livshits and Danny Wyatt and David Esiobu and Dhruv Choudhary and Dhruv Mahajan and Diego Garcia-Olano and Diego Perino and Dieuwke Hupkes and Egor Lakomkin and Ehab AlBadawy and Elina Lobanova and Emily Dinan and Eric Michael Smith and Filip Radenovic and Francisco Guzmán and Frank Zhang and Gabriel Synnaeve and Gabrielle Lee and Georgia Lewis Anderson and Govind Thattai and Graeme Nail and Gregoire Mialon and Guan Pang and Guillem Cucurell and Hailey Nguyen and Hannah Korevaar and Hu Xu and Hugo Touvron and Iliyan Zarov and Imanol Arrieta Ibarra and Isabel Kloumann and Ishan Misra and Ivan Evtimov and Jack Zhang and Jade Copet and Jaewon Lee and Jan Geffert and Jana Vranes and Jason Park and Jay Mahadeokar and Jeet Shah and Jelmer van der Linde and Jennifer Billock and Jenny Hong and Jenya Lee and Jeremy Fu and Jianfeng Chi and Jianyu Huang and Jiawen Liu and Jie Wang and Jiecao Yu and Joanna Bitton and Joe Spisak and Jongsoo Park and Joseph Rocca and Joshua Johnstun and Joshua Saxe and Junteng Jia and Kalyan Vasuden Alwala and Karthik Prasad and Kartikeya Upasani and Kate Plawiak and Ke Li and Kenneth Heafield and Kevin Stone and Khalid El-Arini and Krithika Iyer and Kshitiz Malik and Kuenley Chiu and Kunal Bhalla and Kushal Lakhotia and Lauren Rantala-Yeary and Laurens van der Maaten and Lawrence Chen and Liang Tan and Liz Jenkins and Louis Martin and Lovish Madaan and Lubo Malo and Lukas Blecher and Lukas Landzaat and Luke de Oliveira and Madeline Muzzi and Mahesh Pasupuleti and Mannat Singh and Manohar Paluri and Marcin Kardas and Maria Tsimpoukelli and Mathew Oldham and Mathieu Rita and Maya Pavlova and Melanie Kambadur and Mike Lewis and Min Si and Mitesh Kumar Singh and Mona Hassan and Naman Goyal and Narjes Torabi and Nikolay Bashlykov and Nikolay Bogoychev and Niladri Chatterji and Ning Zhang and Olivier Duchenne and Onur Çelebi and Patrick Alrassy and Pengchuan Zhang and Pengwei Li and Petar Vasic and Peter Weng and Prajjwal Bhargava and Pratik Dubal and Praveen Krishnan and Punit Singh Koura and Puxin Xu and Qing He and Qingxiao Dong and Ragavan Srinivasan and Raj Ganapathy and Ramon Calderer and Ricardo Silveira Cabral and Robert Stojnic and Roberta Raileanu and Rohan Maheswari and Rohit Girdhar and Rohit Patel and Romain Sauvestre and Ronnie Polidoro and Roshan Sumbaly and Ross Taylor and Ruan Silva and Rui Hou and Rui Wang and Saghar Hosseini and Sahana Chennabasappa and Sanjay Singh and Sean Bell and Seohyun Sonia Kim and Sergey Edunov and Shaoliang Nie and Sharan Narang and Sharath Raparthy and Sheng Shen and Shengye Wan and Shruti Bhosale and Shun Zhang and Simon Vandenhende and Soumya Batra and Spencer Whitman and Sten Sootla and Stephane Collot and Suchin Gururangan and Sydney Borodinsky and Tamar Herman and Tara Fowler and Tarek Sheasha and Thomas Georgiou and Thomas Scialom and Tobias Speckbacher and Todor Mihaylov and Tong Xiao and Ujjwal Karn and Vedanuj Goswami and Vibhor Gupta and Vignesh Ramanathan and Viktor Kerkez and Vincent Gonguet and Virginie Do and Vish Vogeti and Vítor Albiero and Vladan Petrovic and Weiwei Chu and Wenhan Xiong and Wenyin Fu and Whitney Meers and Xavier Martinet and Xiaodong Wang and Xiaofang Wang and Xiaoqing Ellen Tan and Xide Xia and Xinfeng Xie and Xuchao Jia and Xuewei Wang and Yaelle Goldschlag and Yashesh Gaur and Yasmine Babaei and Yi Wen and Yiwen Song and Yuchen Zhang and Yue Li and Yuning Mao and Zacharie Delpierre Coudert and Zheng Yan and Zhengxing Chen and Zoe Papakipos and Aaditya Singh and Aayushi Srivastava and Abha Jain and Adam Kelsey and Adam Shajnfeld and Adithya Gangidi and Adolfo Victoria and Ahuva Goldstand and Ajay Menon and Ajay Sharma and Alex Boesenberg and Alexei Baevski and Allie Feinstein and Amanda Kallet and Amit Sangani and Amos Teo and Anam Yunus and Andrei Lupu and Andres Alvarado and Andrew Caples and Andrew Gu and Andrew Ho and Andrew Poulton and Andrew Ryan and Ankit Ramchandani and Annie Dong and Annie Franco and Anuj Goyal and Aparajita Saraf and Arkabandhu Chowdhury and Ashley Gabriel and Ashwin Bharambe and Assaf Eisenman and Azadeh Yazdan and Beau James and Ben Maurer and Benjamin Leonhardi and Bernie Huang and Beth Loyd and Beto De Paola and Bhargavi Paranjape and Bing Liu and Bo Wu and Boyu Ni and Braden Hancock and Bram Wasti and Brandon Spence and Brani Stojkovic and Brian Gamido and Britt Montalvo and Carl Parker and Carly Burton and Catalina Mejia and Ce Liu and Changhan Wang and Changkyu Kim and Chao Zhou and Chester Hu and Ching-Hsiang Chu and Chris Cai and Chris Tindal and Christoph Feichtenhofer and Cynthia Gao and Damon Civin and Dana Beaty and Daniel Kreymer and Daniel Li and David Adkins and David Xu and Davide Testuggine and Delia David and Devi Parikh and Diana Liskovich and Didem Foss and Dingkang Wang and Duc Le and Dustin Holland and Edward Dowling and Eissa Jamil and Elaine Montgomery and Eleonora Presani and Emily Hahn and Emily Wood and Eric-Tuan Le and Erik Brinkman and Esteban Arcaute and Evan Dunbar and Evan Smothers and Fei Sun and Felix Kreuk and Feng Tian and Filippos Kokkinos and Firat Ozgenel and Francesco Caggioni and Frank Kanayet and Frank Seide and Gabriela Medina Florez and Gabriella Schwarz and Gada Badeer and Georgia Swee and Gil Halpern and Grant Herman and Grigory Sizov and Guangyi and Zhang and Guna Lakshminarayanan and Hakan Inan and Hamid Shojanazeri and Han Zou and Hannah Wang and Hanwen Zha and Haroun Habeeb and Harrison Rudolph and Helen Suk and Henry Aspegren and Hunter Goldman and Hongyuan Zhan and Ibrahim Damlaj and Igor Molybog and Igor Tufanov and Ilias Leontiadis and Irina-Elena Veliche and Itai Gat and Jake Weissman and James Geboski and James Kohli and Janice Lam and Japhet Asher and Jean-Baptiste Gaya and Jeff Marcus and Jeff Tang and Jennifer Chan and Jenny Zhen and Jeremy Reizenstein and Jeremy Teboul and Jessica Zhong and Jian Jin and Jingyi Yang and Joe Cummings and Jon Carvill and Jon Shepard and Jonathan McPhie and Jonathan Torres and Josh Ginsburg and Junjie Wang and Kai Wu and Kam Hou U and Karan Saxena and Kartikay Khandelwal and Katayoun Zand and Kathy Matosich and Kaushik Veeraraghavan and Kelly Michelena and Keqian Li and Kiran Jagadeesh and Kun Huang and Kunal Chawla and Kyle Huang and Lailin Chen and Lakshya Garg and Lavender A and Leandro Silva and Lee Bell and Lei Zhang and Liangpeng Guo and Licheng Yu and Liron Moshkovich and Luca Wehrstedt and Madian Khabsa and Manav Avalani and Manish Bhatt and Martynas Mankus and Matan Hasson and Matthew Lennie and Matthias Reso and Maxim Groshev and Maxim Naumov and Maya Lathi and Meghan Keneally and Miao Liu and Michael L. Seltzer and Michal Valko and Michelle Restrepo and Mihir Patel and Mik Vyatskov and Mikayel Samvelyan and Mike Clark and Mike Macey and Mike Wang and Miquel Jubert Hermoso and Mo Metanat and Mohammad Rastegari and Munish Bansal and Nandhini Santhanam and Natascha Parks and Natasha White and Navyata Bawa and Nayan Singhal and Nick Egebo and Nicolas Usunier and Nikhil Mehta and Nikolay Pavlovich Laptev and Ning Dong and Norman Cheng and Oleg Chernoguz and Olivia Hart and Omkar Salpekar and Ozlem Kalinli and Parkin Kent and Parth Parekh and Paul Saab and Pavan Balaji and Pedro Rittner and Philip Bontrager and Pierre Roux and Piotr Dollar and Polina Zvyagina and Prashant Ratanchandani and Pritish Yuvraj and Qian Liang and Rachad Alao and Rachel Rodriguez and Rafi Ayub and Raghotham Murthy and Raghu Nayani and Rahul Mitra and Rangaprabhu Parthasarathy and Raymond Li and Rebekkah Hogan and Robin Battey and Rocky Wang and Russ Howes and Ruty Rinott and Sachin Mehta and Sachin Siby and Sai Jayesh Bondu and Samyak Datta and Sara Chugh and Sara Hunt and Sargun Dhillon and Sasha Sidorov and Satadru Pan and Saurabh Mahajan and Saurabh Verma and Seiji Yamamoto and Sharadh Ramaswamy and Shaun Lindsay and Shaun Lindsay and Sheng Feng and Shenghao Lin and Shengxin Cindy Zha and Shishir Patil and Shiva Shankar and Shuqiang Zhang and Shuqiang Zhang and Sinong Wang and Sneha Agarwal and Soji Sajuyigbe and Soumith Chintala and Stephanie Max and Stephen Chen and Steve Kehoe and Steve Satterfield and Sudarshan Govindaprasad and Sumit Gupta and Summer Deng and Sungmin Cho and Sunny Virk and Suraj Subramanian and Sy Choudhury and Sydney Goldman and Tal Remez and Tamar Glaser and Tamara Best and Thilo Koehler and Thomas Robinson and Tianhe Li and Tianjun Zhang and Tim Matthews and Timothy Chou and Tzook Shaked and Varun Vontimitta and Victoria Ajayi and Victoria Montanez and Vijai Mohan and Vinay Satish Kumar and Vishal Mangla and Vlad Ionescu and Vlad Poenaru and Vlad Tiberiu Mihailescu and Vladimir Ivanov and Wei Li and Wenchen Wang and Wenwen Jiang and Wes Bouaziz and Will Constable and Xiaocheng Tang and Xiaojian Wu and Xiaolan Wang and Xilun Wu and Xinbo Gao and Yaniv Kleinman and Yanjun Chen and Ye Hu and Ye Jia and Ye Qi and Yenda Li and Yilin Zhang and Ying Zhang and Yossi Adi and Youngjin Nam and Yu and Wang and Yu Zhao and Yuchen Hao and Yundi Qian and Yunlu Li and Yuzi He and Zach Rait and Zachary DeVito and Zef Rosnbrick and Zhaoduo Wen and Zhenyu Yang and Zhiwei Zhao and Zhiyu Ma},
      year={2024},
      eprint={2407.21783},
      archivePrefix={arXiv},
      primaryClass={cs.AI},
      url={https://arxiv.org/abs/2407.21783}, 
}

@misc{openai2024gpt4technicalreport,
      title={GPT-4 Technical Report}, 
      author={OpenAI and Josh Achiam and Steven Adler and Sandhini Agarwal and Lama Ahmad and Ilge Akkaya and Florencia Leoni Aleman and Diogo Almeida and Janko Altenschmidt and Sam Altman and Shyamal Anadkat and Red Avila and Igor Babuschkin and Suchir Balaji and Valerie Balcom and Paul Baltescu and Haiming Bao and Mohammad Bavarian and Jeff Belgum and Irwan Bello and Jake Berdine and Gabriel Bernadett-Shapiro and Christopher Berner and Lenny Bogdonoff and Oleg Boiko and Madelaine Boyd and Anna-Luisa Brakman and Greg Brockman and Tim Brooks and Miles Brundage and Kevin Button and Trevor Cai and Rosie Campbell and Andrew Cann and Brittany Carey and Chelsea Carlson and Rory Carmichael and Brooke Chan and Che Chang and Fotis Chantzis and Derek Chen and Sully Chen and Ruby Chen and Jason Chen and Mark Chen and Ben Chess and Chester Cho and Casey Chu and Hyung Won Chung and Dave Cummings and Jeremiah Currier and Yunxing Dai and Cory Decareaux and Thomas Degry and Noah Deutsch and Damien Deville and Arka Dhar and David Dohan and Steve Dowling and Sheila Dunning and Adrien Ecoffet and Atty Eleti and Tyna Eloundou and David Farhi and Liam Fedus and Niko Felix and Simón Posada Fishman and Juston Forte and Isabella Fulford and Leo Gao and Elie Georges and Christian Gibson and Vik Goel and Tarun Gogineni and Gabriel Goh and Rapha Gontijo-Lopes and Jonathan Gordon and Morgan Grafstein and Scott Gray and Ryan Greene and Joshua Gross and Shixiang Shane Gu and Yufei Guo and Chris Hallacy and Jesse Han and Jeff Harris and Yuchen He and Mike Heaton and Johannes Heidecke and Chris Hesse and Alan Hickey and Wade Hickey and Peter Hoeschele and Brandon Houghton and Kenny Hsu and Shengli Hu and Xin Hu and Joost Huizinga and Shantanu Jain and Shawn Jain and Joanne Jang and Angela Jiang and Roger Jiang and Haozhun Jin and Denny Jin and Shino Jomoto and Billie Jonn and Heewoo Jun and Tomer Kaftan and Łukasz Kaiser and Ali Kamali and Ingmar Kanitscheider and Nitish Shirish Keskar and Tabarak Khan and Logan Kilpatrick and Jong Wook Kim and Christina Kim and Yongjik Kim and Jan Hendrik Kirchner and Jamie Kiros and Matt Knight and Daniel Kokotajlo and Łukasz Kondraciuk and Andrew Kondrich and Aris Konstantinidis and Kyle Kosic and Gretchen Krueger and Vishal Kuo and Michael Lampe and Ikai Lan and Teddy Lee and Jan Leike and Jade Leung and Daniel Levy and Chak Ming Li and Rachel Lim and Molly Lin and Stephanie Lin and Mateusz Litwin and Theresa Lopez and Ryan Lowe and Patricia Lue and Anna Makanju and Kim Malfacini and Sam Manning and Todor Markov and Yaniv Markovski and Bianca Martin and Katie Mayer and Andrew Mayne and Bob McGrew and Scott Mayer McKinney and Christine McLeavey and Paul McMillan and Jake McNeil and David Medina and Aalok Mehta and Jacob Menick and Luke Metz and Andrey Mishchenko and Pamela Mishkin and Vinnie Monaco and Evan Morikawa and Daniel Mossing and Tong Mu and Mira Murati and Oleg Murk and David Mély and Ashvin Nair and Reiichiro Nakano and Rajeev Nayak and Arvind Neelakantan and Richard Ngo and Hyeonwoo Noh and Long Ouyang and Cullen O'Keefe and Jakub Pachocki and Alex Paino and Joe Palermo and Ashley Pantuliano and Giambattista Parascandolo and Joel Parish and Emy Parparita and Alex Passos and Mikhail Pavlov and Andrew Peng and Adam Perelman and Filipe de Avila Belbute Peres and Michael Petrov and Henrique Ponde de Oliveira Pinto and Michael and Pokorny and Michelle Pokrass and Vitchyr H. Pong and Tolly Powell and Alethea Power and Boris Power and Elizabeth Proehl and Raul Puri and Alec Radford and Jack Rae and Aditya Ramesh and Cameron Raymond and Francis Real and Kendra Rimbach and Carl Ross and Bob Rotsted and Henri Roussez and Nick Ryder and Mario Saltarelli and Ted Sanders and Shibani Santurkar and Girish Sastry and Heather Schmidt and David Schnurr and John Schulman and Daniel Selsam and Kyla Sheppard and Toki Sherbakov and Jessica Shieh and Sarah Shoker and Pranav Shyam and Szymon Sidor and Eric Sigler and Maddie Simens and Jordan Sitkin and Katarina Slama and Ian Sohl and Benjamin Sokolowsky and Yang Song and Natalie Staudacher and Felipe Petroski Such and Natalie Summers and Ilya Sutskever and Jie Tang and Nikolas Tezak and Madeleine B. Thompson and Phil Tillet and Amin Tootoonchian and Elizabeth Tseng and Preston Tuggle and Nick Turley and Jerry Tworek and Juan Felipe Cerón Uribe and Andrea Vallone and Arun Vijayvergiya and Chelsea Voss and Carroll Wainwright and Justin Jay Wang and Alvin Wang and Ben Wang and Jonathan Ward and Jason Wei and CJ Weinmann and Akila Welihinda and Peter Welinder and Jiayi Weng and Lilian Weng and Matt Wiethoff and Dave Willner and Clemens Winter and Samuel Wolrich and Hannah Wong and Lauren Workman and Sherwin Wu and Jeff Wu and Michael Wu and Kai Xiao and Tao Xu and Sarah Yoo and Kevin Yu and Qiming Yuan and Wojciech Zaremba and Rowan Zellers and Chong Zhang and Marvin Zhang and Shengjia Zhao and Tianhao Zheng and Juntang Zhuang and William Zhuk and Barret Zoph},
      year={2024},
      eprint={2303.08774},
      archivePrefix={arXiv},
      primaryClass={cs.CL},
      url={https://arxiv.org/abs/2303.08774}, 
}

@inproceedings{han2025token,
  title={Token-budget-aware llm reasoning},
  author={Han, Tingxu and Wang, Zhenting and Fang, Chunrong and Zhao, Shiyu and Ma, Shiqing and Chen, Zhenyu},
  booktitle={Findings of the Association for Computational Linguistics: ACL 2025},
  pages={24842--24855},
  year={2025}
}

@article{ma2025reasoning,
  title={Reasoning models can be effective without thinking},
  author={Ma, Wenjie and He, Jingxuan and Snell, Charlie and Griggs, Tyler and Min, Sewon and Zaharia, Matei},
  journal={arXiv preprint arXiv:2504.09858},
  year={2025}
}

@inproceedings{
frantar2023optq,
title={{OPTQ}: Accurate Quantization for Generative Pre-trained Transformers},
author={Elias Frantar and Saleh Ashkboos and Torsten Hoefler and Dan Alistarh},
booktitle={The Eleventh International Conference on Learning Representations },
year={2023},
url={https://openreview.net/forum?id=tcbBPnfwxS}
}

@inproceedings{yu2025thinkrec,
  title={ThinkRec: Thinking-based recommendation via LLM},
  author={Yu, Qihang and Fu, Kairui and Zhang, Shengyu and Lv, Zheqi and Wu, Fan and Wu, Fei},
  booktitle={Proceedings of the ACM Web Conference 2026},
  year={2026}
}

@inproceedings{lv2025collaboration,
  title={Collaboration of Large Language Models and Small Recommendation Models for Device-Cloud Recommendation},
  author={Lv, Zheqi and Zhan, Tianyu and Wang, Wenjie and Lin, Xinyu and Zhang, Shengyu and Zhang, Wenqiao and Li, Jiwei and Kuang, Kun and Wu, Fei},
  booktitle={Proceedings of the 31st ACM SIGKDD Conference on Knowledge Discovery and Data Mining V. 1},
  pages={962--973},
  year={2025}
}

@article{lin2025healthgpt,
  title={Healthgpt: A medical large vision-language model for unifying comprehension and generation via heterogeneous knowledge adaptation},
  author={Lin, Tianwei and Zhang, Wenqiao and Li, Sijing and Yuan, Yuqian and Yu, Binhe and Li, Haoyuan and He, Wanggui and Jiang, Hao and Li, Mengze and Song, Xiaohui and others},
  journal={arXiv preprint arXiv:2502.09838},
  year={2025}
}

@inproceedings{yuan2025videorefer,
  title={Videorefer suite: Advancing spatial-temporal object understanding with video llm},
  author={Yuan, Yuqian and Zhang, Hang and Li, Wentong and Cheng, Zesen and Zhang, Boqiang and Li, Long and Li, Xin and Zhao, Deli and Zhang, Wenqiao and Zhuang, Yueting and others},
  booktitle={Proceedings of the Computer Vision and Pattern Recognition Conference},
  pages={18970--18980},
  year={2025}
}

@inproceedings{zhang2022boostmis,
  title={Boostmis: Boosting medical image semi-supervised learning with adaptive pseudo labeling and informative active annotation},
  author={Zhang, Wenqiao and Zhu, Lei and Hallinan, James and Zhang, Shengyu and Makmur, Andrew and Cai, Qingpeng and Ooi, Beng Chin},
  booktitle={Proceedings of the IEEE/CVF Conference on Computer Vision and Pattern Recognition},
  pages={20666--20676},
  year={2022}
}

@article{brown2020language,
  title={Language models are few-shot learners},
  author={Brown, Tom and Mann, Benjamin and Ryder, Nick and Subbiah, Melanie and Kaplan, Jared D and Dhariwal, Prafulla and Neelakantan, Arvind and Shyam, Pranav and Sastry, Girish and Askell, Amanda and others},
  journal={Advances in neural information processing systems},
  volume={33},
  pages={1877--1901},
  year={2020}
}

\appendix
\clearpage
% \section{Example Appendix}
\section{Implementation Details}

\label{sec:appendix}

\paragraph{Details of HRA.}
Tables~\ref{tab:gsm8k_cot_length} and~\ref{tab:math_cot_length} report the average CoT lengths of the four \emph{Hierarchical Reasoning Abstraction} tiers (\textit{Detailed}, \textit{Standard}, \textit{Concise}, \textit{Ultra-Concise}) on GSM8K and MATH-500. Across both datasets, CoT length decreases consistently with the abstraction tier, highlighting the systematic impact of hierarchical abstraction on reasoning verbosity.

\begin{table}[htbp]
\centering
\small
\begin{tabular}{lcccc}
\toprule
\textbf{Model} & \textbf{D} & \textbf{S} & \textbf{C} & \textbf{UC} \\
\midrule
Llama3.1-8B    & 269.24 & 215.13 & 142.54 & 106.67 \\
Qwen2.5-3B     & 369.32 & 368.52 & 183.02 & 99.78  \\
Qwen2.5-7B     & 268.48 & 237.19 & 166.23 & 121.76 \\
Qwen2.5-14B    & 296.92 & 254.20 & 219.94 & 152.04 \\
\bottomrule
\end{tabular}
\caption{Average CoT length of different models on the GSM8K dataset
under different verbosity settings (D: Detailed, S: Standard, C: Concise, UC: Ultra-Concise).}
\label{tab:gsm8k_cot_length}
\end{table}

\begin{table}[htbp]
\centering
\small
\begin{tabular}{lcccc}
\toprule
\textbf{Model} & \textbf{D} & \textbf{S} & \textbf{C} & \textbf{UC} \\
\midrule
Llama3.1-8B     & 510.07 & 432.58 & 285.67 & 172.48 \\
Qwen2.5-3B      & 748.87 & 721.97 & 527.64 & 219.62 \\
Qwen2.5-7B      & 601.39 & 525.56 & 461.95 & 258.21 \\
Qwen2.5-14B     & 625.97 & 566.12 & 397.19 & 235.89 \\
\bottomrule
\end{tabular}
\caption{Average CoT length of different models on the MATH-500 dataset
under different verbosity settings.}
\label{tab:math_cot_length}
\end{table}

\begin{table}[H]
\centering
\small
\setlength{\tabcolsep}{6pt}
\begin{tabular}{l l}
\toprule
\textbf{Hyperparameter} & \textbf{Value} \\
\midrule
LoRA rank $r$ & 8 \\
LoRA alpha $\alpha$ & 16 \\
\midrule
Optimizer & AdamW \\
Learning rate & $5\times 10^{-5}$ \\
LR scheduler & cosine \\
Warmup ratio & 0.1 \\
Training epochs & 3 \\
\midrule
Per-device train batch size & 1 \\
Gradient accumulation steps & 8 \\
Precision & BF16 \\
\bottomrule
\end{tabular}
\caption{Key training hyperparameters (shared across backbones).}
\label{tab:train_hparams}
\end{table}

\paragraph{Training setup.}
% We fine-tune Qwen2.5-Instruct and Llama3-Instruct using LLaMA-Factory with supervised fine-tuning (SFT).
% We adopt LoRA and apply it to all target modules (\texttt{lora\_target=all}).
% % The training data is our compressed math instruction dataset
% % (\texttt{mydataset\_compressed\_math\_llmlingua2\_qwen2.5\_7b\_tokenskip\_math\_1024\_1027}) formatted with the \texttt{qwen} template.

% % Distributed training uses \texttt{ddp\_timeout=180000000}.
% % We log training statistics every 10 steps and save checkpoints every 300 steps.
% % For evaluation, we hold out 10\% of the training set as validation (\texttt{val\_size=0.1}) and evaluate every 300 steps (\texttt{eval\_strategy=steps}, \texttt{eval\_steps=300}) with per-device evaluation batch size 1.

% \begin{table}[t]
% \centering
% \small
% \setlength{\tabcolsep}{6pt}
% \begin{tabular}{l l}
% \toprule
% \textbf{Hyperparameter} & \textbf{Value} \\
% \midrule
% LoRA rank $r$ & 8 \\
% LoRA alpha $\alpha$ & 16 \\
% \midrule
% % Max sequence length (\texttt{cutoff\_len}) & 2048 \\
% % Max training samples (\texttt{max\_samples}) & 10,000 \\
% % Preprocess workers & 16 \\
% % \midrule
% Optimizer & AdamW \\
% Learning rate & $5\times 10^{-5}$ \\
% LR scheduler & cosine \\
% Warmup ratio & 0.1 \\
% Training epochs & 3 \\
% \midrule
% Per-device train batch size & 1 \\
% Gradient accumulation steps & 8 \\
% Precision & BF16 \\
% \bottomrule
% \end{tabular}
% \caption{Key training hyperparameters (shared across backbones).}
% \label{tab:train_hparams}
% \end{table}

We conduct LoRA-based supervised fine-tuning (SFT) using LLaMA-Factory, and apply LoRA to all target modules.
% All experiments are run on a single machine equipped with $8\times$ NVIDIA RTX 4090D GPUs, an Intel(R) Xeon(R) Platinum 8474C CPU, CUDA 12.4, and PyTorch 2.5.1.
The Qwen2.5-14B-Instruct experiments are run on NVIDIA A800 GPUs with an Intel(R) Xeon(R) Gold 6348 CPU, while all other experiments are run on NVIDIA RTX 4090D GPUs with an Intel(R) Xeon(R) Platinum 8474C CPU; all runs use CUDA 12.4 and PyTorch 2.5.1.

The key training hyperparameters are summarized in Table~\ref{tab:train_hparams}.

\paragraph{Dataset size and training time.}
Table~\ref{tab:data_time} reports the number of training instances and the end-to-end wall-clock training time for model and dataset.

\begin{table}[H]
\centering
\small
\setlength{\tabcolsep}{6pt}
\begin{tabular}{l l r r}
\toprule
\textbf{Model} & \textbf{Dataset} & \textbf{Samples} & \textbf{Time} \\
\midrule
Qwen2.5-3B  & MATH-500  & 17{,}179 & 03:45:01 \\
Qwen2.5-3B  & GSM8K & 22{,}813 & 04:08:41 \\
Qwen2.5-7B  & MATH-500  & 28{,}062 & 06:57:10 \\
Qwen2.5-7B  & GSM8K & 31{,}125 & 04:47:33 \\
Qwen2.5-14B & MATH-500  & 21{,}780 & 07:32:33 \\
Qwen2.5-14B & GSM8K & 33{,}925 & 09:40:38 \\
Llama3.1-8B & MATH-500  & 27{,}660 & 07:10:46 \\
Llama3.1-8B & GSM8K & 33{,}678 & 08:22:42 \\
\bottomrule
\end{tabular}
\caption{Training set sizes and wall-clock training time for LoRA-SFT across backbones and datasets.}
\label{tab:data_time}
\end{table}

\begin{figure}[H]       % [t]=顶部 [h]=当前位置 [b]=底部 [!ht]=尽量当前位置
  \centering
  \includegraphics[width=\linewidth]{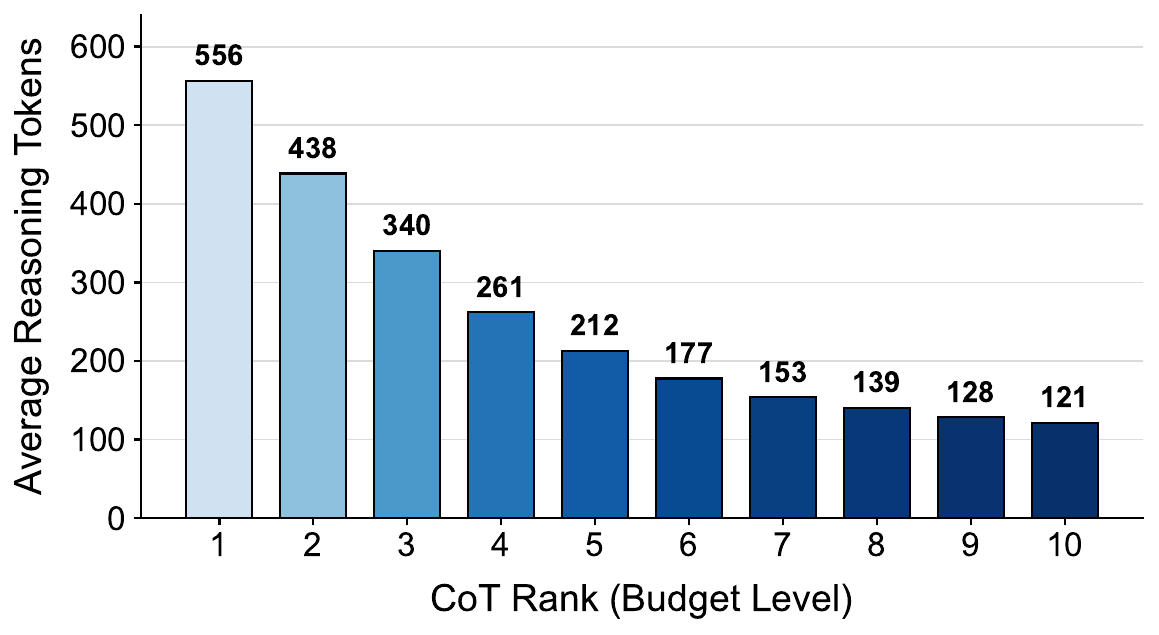} % 也可以写成 width=0.8\linewidth
  \vspace{-0.5cm}
\caption{Average CoT length (reasoning tokens) across budget levels on \textsc{MATH-500} with Qwen2.5-7B-Instruct. CoT rank (1--10) denotes increasingly tighter budgets; bars report the mean length of the corresponding CoTs produced by our method.}

  \label{fig:xxx}
\vspace{-0.3cm}
\end{figure}

\paragraph{CoT Length Analysis.}
Figure~\ref{fig:xxx} summarizes the average length of the CoTs produced by our method on \textsc{MATH-500} using Qwen2.5-7B-Instruct.
As the budget level tightens (CoT rank 1 $\rightarrow$ 10), the mean CoT length decreases monotonically from 556 to 121 reasoning tokens (a $\sim$4.6$\times$ reduction), forming a smooth spectrum of trace lengths.
This trend indicates that the proposed budget control provides stable, fine-grained length modulation for mathematical reasoning.

\begin{table*}[t]
\centering
\small
\setlength{\tabcolsep}{7pt}
\renewcommand{\arraystretch}{0.95}
\resizebox{\linewidth}{!}{
\begin{tabular}{l l l | r r r r | r r r r}
\toprule[1.5pt]
% \multirow{2}{*}{\multicolumn{1}{c}{\textbf{Model}}} &
\multirow{2}{*}{\textbf{Model}} &
% \multirow{2}{*}{\multicolumn{1}{c}{\textbf{CS.}}} &
% \multirow{2}{*}{\multicolumn{1}{c}{\textbf{Methods}}} &
\multirow{2}{*}{\textbf{CS.}} &
\multirow{2}{*}{\textbf{Methods}} &
\multicolumn{4}{c|}{\textbf{GSM8K}} &
\multicolumn{4}{c}{\textbf{MATH-500}} \\
\cmidrule(lr){4-7}\cmidrule(lr){8-11}
& & &
\textbf{Acc. $\uparrow$} &
\textbf{Tokens $\downarrow$} &
\textbf{CR. $\downarrow$} &
\textbf{TE. $\uparrow$} &
\textbf{Acc. $\uparrow$} &
\textbf{Tokens $\downarrow$} &
\textbf{CR. $\downarrow$} &
\textbf{TE. $\uparrow$} \\
\midrule

% ================= Qwen2.5-7B =================
\multirow{7}{*}{\textbf{Qwen2.5-7B}}
& --- & Original &
91.58 & 299.22 & 1.00 & 3.27 &
71.20 & 574.58 & 1.00 & 7.70 \\
\cmidrule(lr){2-11}

% & \multirow{4}{*}{Low$^{-}$} & \textcolor{black!60}{LC\text{-}Prompt} & \textcolor{black!60}{91.58} & \textcolor{black!60}{268.69} & \textcolor{black!60}{0.90} & \textcolor{black!60}{34.09} & \textcolor{black!60}{68.20} & \textcolor{black!60}{555.97} & \textcolor{black!60}{0.97} & \textcolor{black!60}{12.27} \\
& \multirow{3}{*}{Low$^{-}$} & Truncation  & 85.97 & 292.77 & 0.98 & 29.37 & 65.80 & 544.86 & 0.95 & 12.08 \\
&  & TokenSkip & \textbf{90.14} & 243.77 & 0.81 & 36.98 & 67.60 & 491.76 & 0.86 & 13.75 \\
&  & \textbf{Ours}& 89.31 & \textbf{218.22} & \textbf{0.73} & \textbf{40.93} & \textbf{68.80} & \textbf{428.68} & \textbf{0.75} & \textbf{16.05} \\
\cmidrule(lr){2-11}
% & \multirow{4}{*}{Low$^{--}$} & \textcolor{black!60}{LC\text{-}Prompt} & \textcolor{black!60}{90.60} & \textcolor{black!60}{266.16} & \textcolor{black!60}{0.89} & \textcolor{black!60}{34.04} & \textcolor{black!60}{69.40} & \textcolor{black!60}{553.90} & \textcolor{black!60}{0.96} & \textcolor{black!60}{12.53} \\
& \multirow{3}{*}{Low$^{--}$} & Truncation  & 90.07 & 296.73 & 0.99 & 30.35 & 69.00 & 563.32 & 0.98 & 12.25 \\
&  & TokenSkip  & 90.60 & 262.53 & 0.88 & 34.51 & 69.60 & 519.78 & 0.90 & 13.39 \\
&  & \textbf{Ours} & \textbf{91.13} & \textbf{252.92} & \textbf{0.85} & \textbf{36.03} & \textbf{70.60} & \textbf{449.46} & \textbf{0.78} & \textbf{15.71} \\
% \cmidrule(lr){2-11}
\midrule
% ================= Qwen2.5-14B =================
\multirow{7}{*}{\textbf{Qwen2.5-14B}}
& --- & Original
& 93.03 & 313.94 & 1.00 & 29.63
& 75.80 & 583.66 & 1.00 & 12.99 \\
\cmidrule(lr){2-11}
% & \multirow{4}{*}{Low$^{-}$} & \textcolor{black!60}{LC\text{-}Prompt} & \textcolor{black!60}{93.48} & \textcolor{black!60}{288.70} & \textcolor{black!60}{0.92} & \textcolor{black!60}{32.38} & \textcolor{black!60}{76.20} & \textcolor{black!60}{559.07} & \textcolor{black!60}{0.96} & \textcolor{black!60}{13.63} \\
& \multirow{3}{*}{Low$^{-}$} & Truncation  & 86.20 & 305.95 & 0.97 & 28.17 & 69.40 & 552.18 & 0.95 & 12.57 \\
&  & TokenSkip & 93.48 & 250.83 & 0.80 & 37.27 & 72.20 & 474.31 & 0.81 & 15.22 \\
&  & \textbf{Ours} & \textbf{93.63} & \textbf{243.90} & \textbf{0.78} & \textbf{38.39} & \textbf{75.40} & \textbf{457.45} & \textbf{0.78} & \textbf{16.48} \\
\cmidrule(lr){2-11}
% & \multirow{4}{*}{Low$^{--}$} & \textcolor{black!60}{LC\text{-}Prompt} & \textcolor{black!60}{93.03} & \textcolor{black!60}{292.28} & \textcolor{black!60}{0.93} & \textcolor{black!60}{31.83} & \textcolor{black!60}{75.80} & \textcolor{black!60}{565.95} & \textcolor{black!60}{0.97} & \textcolor{black!60}{13.39} \\
& \multirow{3}{*}{Low$^{--}$} & Truncation & 90.30 & 311.37 & 0.99 & 29.00 & 73.60 & 570.48 & 0.98 & 12.90 \\
&  & TokenSkip  & \textbf{94.16} & 269.52 & 0.86 & 34.94 & 73.40 & 496.55 & 0.85 & 14.78 \\
&  & \textbf{Ours} & 94.01 & \textbf{261.18} & \textbf{0.83} & \textbf{35.99} & \textbf{74.80} & \textbf{481.94} & \textbf{0.83} & \textbf{15.52} \\
\bottomrule[1.5pt]
\end{tabular}
}
\caption{Additional results under milder compression strengths (Low$^{-}$/$^{--}$) on GSM8K and MATH-500.}
\label{tab:more_result}
\end{table*}

\begin{table*}[t]
\centering
\small
\setlength{\tabcolsep}{7pt}
\renewcommand{\arraystretch}{0.95}
\resizebox{\linewidth}{!}{
\begin{tabular}{l l l | r r r r | r r r r}
\toprule[1.5pt]
% \multirow{2}{*}{\multicolumn{1}{c}{\textbf{Model}}} &
\multirow{2}{*}{\textbf{Model}} &
% \multirow{2}{*}{\multicolumn{1}{c}{\textbf{CS.}}} &
% \multirow{2}{*}{\multicolumn{1}{c}{\textbf{Methods}}} &
\multirow{2}{*}{\textbf{CS.}} &
\multirow{2}{*}{\textbf{Methods}} &
\multicolumn{4}{c|}{\textbf{GSM8K}} &
\multicolumn{4}{c}{\textbf{MATH-500}} \\
\cmidrule(lr){4-7}\cmidrule(lr){8-11}
& & &
\textbf{Acc. $\uparrow$} &
\textbf{Tokens $\downarrow$} &
\textbf{CR. $\downarrow$} &
\textbf{TE. $\uparrow$} &
\textbf{Acc. $\uparrow$} &
\textbf{Tokens $\downarrow$} &
\textbf{CR. $\downarrow$} &
\textbf{TE. $\uparrow$} \\
\midrule
% ================= Qwen2.5-3B =================
\multirow{10}{*}{\textbf{Qwen2.5-3B}}
& --- & Original &
83.24 & 316.94 & 1.00 & 3.81 &
63.20 & 575.66 & 1.00 & 9.11 \\
\cmidrule(lr){2-11}
& \multirow{3}{*}{High$^{+++}$} & Truncation & 2.88 & 102.00 & 0.32 & 2.82 & 6.40 & 185.11 & 0.32 & 3.46 \\
& & TokenSkip & 48.07 & 80.70 & 0.25 & 59.56 & 18.60 & 167.11 & 0.29 & 11.13 \\
&  & Ours     & \textbf{63.38} & \textbf{74.40} & \textbf{0.23} & \textbf{85.19} & \textbf{34.80} & \textbf{140.06} & \textbf{0.24} & \textbf{24.85} \\
\cmidrule(lr){2-11} 
& \multirow{3}{*}{High$^{++}$} & Truncation & 6.14 & 153.92 & 0.49 & 3.99 & 18.20 & 299.73 & 0.52 & 6.07 \\
&  & TokenSkip & 58.68 & 108.07 & 0.34 & 54.30 & 22.80 & \textbf{222.00} & \textbf{0.39} & 10.27 \\
&  & Ours     & \textbf{74.30} & \textbf{106.24} & \textbf{0.34} & \textbf{69.93} & \textbf{42.00} & 245.06 & 0.43 & \textbf{17.14} \\
\cmidrule(lr){2-11}
& \multirow{3}{*}{High$^{+}$} & Truncation & 18.35 & 202.69 & 0.64 & 9.05  & 33.00 & 377.83 & 0.66 & 8.73 \\
&  & TokenSkip & 68.01 & 134.17 & 0.42 & 50.69 & 31.40 & 284.91 & 0.49 & 11.02 \\
&  & Ours     & \textbf{77.10} & \textbf{125.44} & \textbf{0.40} & \textbf{61.47} & \textbf{39.80} & \textbf{249.26} & \textbf{0.43} & \textbf{15.97} \\
\midrule
% ================= Qwen2.5-7B =================
\multirow{10}{*}{\textbf{Qwen2.5-7B}}
& --- & Original &
91.58 & 299.22 & 1.00 & 3.27 &
71.20 & 574.58 & 1.00 & 7.70 \\
\cmidrule(lr){2-11}
& \multirow{2}{*}{High$^{+++}$} & Truncation & 2.58 & 102.00 & 0.34 & 2.53 & 5.00 & 165.23 & 0.29 & 3.03 \\
&  & TokenSkip & 61.18 & 66.95 & 0.22 & 91.38 & 28.20 & 142.76 & 0.25 & 19.75 \\
&  & Ours     & \textbf{73.69} & \textbf{58.83} & \textbf{0.20} & \textbf{125.27} & \textbf{49.40} & \textbf{140.17} & \textbf{0.24} & \textbf{35.24} \\
\cmidrule(lr){2-11} 
& \multirow{2}{*}{High$^{++}$} & Truncation & 6.75 & 153.81 & 0.51 & 4.39 & 18.00 & 300.29 & 0.52 & 5.99 \\
&  & TokenSkip & 72.40 & 93.01 & 0.31 & 77.85 & 35.80 & 199.87 & 0.35 & 17.91 \\
&  & Ours     & \textbf{79.08} & \textbf{83.36} & \textbf{0.28} & \textbf{94.86} & \textbf{54.00} & \textbf{187.34} & \textbf{0.33} & \textbf{28.82} \\
\cmidrule(lr){2-11} 
& \multirow{2}{*}{High$^{+}$}  & Truncation & 23.28 & 201.79 & 0.67 & 11.54 & 36.60 & 379.45 & 0.66 & 9.65 \\
&  & TokenSkip & 79.30 & 130.45 & 0.44 & 60.79 & 43.60 & 258.58 & 0.45 & 16.86 \\
&  & Ours     & \textbf{83.17} & \textbf{109.80} & \textbf{0.37} & \textbf{75.74} & \textbf{58.00} & \textbf{225.61} & \textbf{0.39} & \textbf{25.71} \\
\midrule
% ================= Qwen2.5-14B =================
\multirow{10}{*}{\textbf{Qwen2.5-14B}}
& --- & Original
& 93.03 & 313.94 & 1.00 & 29.63
& 75.80 & 583.66 & 1.00 & 12.99 \\
\cmidrule(lr){2-11}
& \multirow{2}{*}{High$^{+++}$} & Truncation& 2.05 & 102.00 & 0.32 & 2.01 & 5.40 & 204.71 & 0.35 & 2.64 \\
&  & TokenSkip & 77.18 & 75.95 & 0.24 & 101.62 & 33.20 & 164.98 & 0.28 & 20.12 \\
&  & Ours     & \textbf{79.83} & \textbf{67.97} & \textbf{0.22} & \textbf{117.44} & \textbf{58.00} & \textbf{137.55} & \textbf{0.24} & \textbf{42.17} \\
\cmidrule(lr){2-11} 
& \multirow{2}{*}{High$^{++}$} & Truncation & 5.69 & 153.93 & 0.49 & 3.69 & 16.60 & 301.09 & 0.52 & 5.51 \\
&  & TokenSkip & 84.46 & 106.01 & 0.34 & 79.67 & 44.40 & 227.86 & 0.39 & 19.49 \\
&  & Ours     & \textbf{87.49} & \textbf{100.41} & \textbf{0.32} & \textbf{87.13} & \textbf{60.20} & \textbf{188.61} & \textbf{0.32} & \textbf{31.92} \\
\cmidrule(lr){2-11} 
& \multirow{2}{*}{High$^{+}$}  & Truncation & 18.35 & 203.00 & 0.65 & 9.04 & 33.60 & 381.41 & 0.65 & 8.81 \\
&  & TokenSkip & 88.40 & 135.74 & 0.43 & 65.12 & 54.80 & 285.42 & 0.49 & 19.20 \\
&  & Ours     & \textbf{90.14} & \textbf{120.92} & \textbf{0.39} & \textbf{74.55} & \textbf{66.60} & \textbf{270.80} & \textbf{0.46} & \textbf{24.59} \\
\bottomrule[1.5pt]
\end{tabular}
}
\caption{Additional results under \emph{stronger} compression strengths (High$^{+}$/$^{++}$/$^{+++}$) on GSM8K and MATH-500.}
\label{tab:more_aggressive}
\end{table*}

% \subsection{CoT Recovery from Compressed Inputs}
\section{Additional Experimental Results}
\paragraph{Milder Compression.}
In the main paper, we report results at Low/Mid/High compression strengths.
Table~\ref{tab:more_result} further presents two milder settings, Low$^{-}$ and Low$^{--}$, on Qwen2.5-7B/14B.
Overall, our method remains robust when compression is more conservative: it consistently reduces CoT length while largely preserving accuracy.
For example, under Low$^{-}$ compression on MATH-500 with Qwen2.5-7B, our method uses over 60 fewer tokens than TokenSkip while achieving higher accuracy. 
On Qwen2.5-14B, our method similarly achieves higher or comparable accuracy using fewer tokens throughout.
By contrast, baselines either provide limited compression (e.g., Truncation) or exhibit a worse compression--accuracy trade-off (e.g., TokenSkip).
These results complement the main-table findings and confirm that our approach delivers stable improvements across a wide range of compression strengths.

\paragraph{More Aggressive Compression.}
Table~\ref{tab:more_aggressive} reports results under stronger compression (High$^{+}$/$^{++}$/$^{+++}$), where the budgets are much tighter than those used in the main paper.
Across all model scales, TokenSkip suffers a sharp accuracy degradation when pushed to these regimes.
In contrast, our method remains substantially more robust: it consistently achieves much higher accuracy while using comparable or fewer tokens.

For instance, on Qwen2.5-7B with High$^{+++}$, TokenSkip drops to 48.07/18.60 accuracy on GSM8K/MATH-500, whereas ours reaches 63.38/34.80 with fewer tokens on both datasets (74.40 vs.\ 80.70; 140.06 vs.\ 167.11).
Similar trends hold for Qwen2.5-3B and Qwen2.5-14B, where our method maintains markedly better accuracy at comparable compression ratios.
Overall, these results show that our dual-granularity approach is particularly beneficial in the most budget-constrained settings, where token-only skipping becomes brittle.

\paragraph{Compression Strength Details.}
Table~\ref{tab:budgets} summarizes the compression-strength (CS) configurations used for LC-Prompt, Truncation, and TokenSkip.
CS levels from high$^{+++}$ to low$^{--}$ are mapped to fixed compression ratios (CR) spanning 0.2 (strongest compression) to 0.9 (weakest compression), providing a unified control knob for varying compression intensity and ensuring fair cross-method comparison under matched CR settings.
Table~\ref{tab:budgets_token} reports the token budgets used for budget-controlled evaluation across model scales and datasets.
Budgets are selected per model--dataset pair to yield compact traces without overly aggressive truncation; accordingly, harder benchmarks (e.g., MATH-500) generally adopt larger budgets.

\begin{table}[t]
\centering
\small
\setlength{\tabcolsep}{8pt}
\begin{tabular}{l r}
\toprule
\textbf{CS.} & \textbf{CR} \\
\midrule
High$^{+++}$ & 0.2 \\
High$^{++}$  & 0.3 \\
High$^{+}$   & 0.4 \\
High         & 0.5 \\
Mid          & 0.6 \\
Low          & 0.7 \\
Low$^{-}$    & 0.8 \\
Low$^{--}$   & 0.9 \\
\bottomrule
\end{tabular}
\caption{Compression ratios (CR) corresponding to different control settings (CS.) from High$^{+++}$ to Low$^{--}$.}
\label{tab:budgets}
\end{table}

\begin{table}[h]
\centering
\small
\setlength{\tabcolsep}{6pt}
\begin{tabular}{l l r r}
\toprule
\textbf{CS.} & \textbf{Model} & \textbf{GSM8K} & \textbf{MATH-500} \\
\midrule
\multirow{4}{*}{\centering High}
  & Llama3.1-8B  & 50  & 100 \\
  & Qwen2.5-3B   & 125 & 200 \\
  & Qwen2.5-7B   & 125 & 150 \\
  & Qwen2.5-14B  & 125 & 200 \\
\midrule
\multirow{4}{*}{\centering Mid}
  & Llama3.1-8B  & 75  & 150 \\
  & Qwen2.5-3B   & 150 & 250 \\
  & Qwen2.5-7B   & 150 & 200 \\
  & Qwen2.5-14B  & 175 & 250 \\
\midrule
\multirow{4}{*}{\centering Low}
  & Llama3.1-8B  & 100 & 200 \\
  & Qwen2.5-3B   & 200 & 300 \\
  & Qwen2.5-7B   & 175 & 400 \\
  & Qwen2.5-14B  & 200 & 350 \\
\midrule
\multirow{2}{*}{\centering Low$^{-}$}
  & Qwen2.5-7B   & 200 & 400 \\
  & Qwen2.5-14B  & 175 & 400 \\
\midrule
\multirow{2}{*}{\centering Low$^{--}$}
  & Qwen2.5-7B   & 250 & 450 \\
  & Qwen2.5-14B  & 200 & 450 \\
\bottomrule
\end{tabular}
\caption{Token budgets (CS.) used for budget-controlled generation on GSM8K and MATH-500 across different models.}
\label{tab:budgets_token}
\end{table}

\section{Distillation of Token Pruner}
\label{sec:distill_pruner}
\paragraph{Data Collection.}
We follow the training pipeline of LLMLingua2~\citep{pan2024llmlingua} and only change the supervision source.
Concretely, we run Qwen2.5-14B-Instruct on the MATH-500 training split to generate a CoT and a final answer for each problem.
We keep the instances whose predicted answers exactly match the ground truth, and randomly sample 1{,}000 correct CoT traces to form $\{r^{(k)}\}_{k=1}^{1000}$.

\paragraph{Teacher Compression and Token Labeling.}
For each correct CoT $r^{(k)}$, we query \textsc{GPT-4} to obtain a compressed CoT trace $\bar{r}^{(k)}$.
We then align the tokenized sequences of $r^{(k)}$ and $\bar{r}^{(k)}$ to derive token-level supervision.
Specifically, each token in $r^{(k)}$ is labeled as \texttt{true} if it is selected by the alignment, and \texttt{false} otherwise, yielding a boolean label sequence
$y^{(k)}=(y^{(k)}_1,\ldots,y^{(k)}_{n_k})$ with $y^{(k)}_i\in\{\texttt{true},\texttt{false}\}$.
Aggregating across all traces gives a labeled dataset $\{(x^{(k)}, y^{(k)})\}_{k=1}^{K}$, where $x^{(k)}=(x^{(k)}_1,\ldots,x^{(k)}_{n_k})$ denotes the token sequence of $r^{(k)}$.

\paragraph{Pruner Architecture.}
Following LLMLingua2, we model token pruning as token-level classification.
Given a token sequence $x=(x_1,\ldots,x_n)$, a bidirectional Transformer encoder produces contextual representations:
\begin{equation}
\mathbf{h}=f_{\theta}(x), \qquad \mathbf{h}=(h_1,\ldots,h_n),
\end{equation}
where $h_i$ is the contextual embedding of token $x_i$.
A lightweight classification head maps each $h_i$ to a keep/drop distribution:
\begin{equation}
p(x_i;\Theta)=\mathrm{softmax}(W h_i + b),
\end{equation}
where $\Theta=\{\theta,W,b\}$ and $p(x_i;\Theta)\in\mathbb{R}^2$ corresponds to the probabilities of \texttt{true} (keep) and \texttt{false} (drop).

\paragraph{Training Objective.}
The pruner is trained with token-level cross-entropy supervision.
Let $y_i\in\{\texttt{true},\texttt{false}\}$ denote the keep/drop label and $p(x_i;\Theta)$ the predicted distribution.
We minimize the average cross-entropy loss over all labeled tokens:
\begin{equation}
\mathcal{L}(\Theta)=\frac{1}{N}\sum_{i=1}^{N}\mathrm{CrossEntropy}\bigl(y_i,\,p(x_i;\Theta)\bigr).
\end{equation}
After 10 epochs of training, the pruner achieves higher pruning accuracy on math reasoning traces, yielding more reliable keep/drop decisions for mathematical CoTs.

\begin{table*}[t]
\centering
\small
\setlength{\tabcolsep}{6pt}
\renewcommand{\arraystretch}{1.15}
\begin{tabularx}{\textwidth}{p{0.22\textwidth} X}
\toprule
\textbf{Level} & \textbf{Prompt Template} \\
\midrule

Detailed CoT &
\begin{minipage}[t]{\linewidth}\vspace{0pt}\ttfamily\footnotesize
You are a mathematics expert. Now I will present you with a mathematical problem and its correct answer. You need to output the correct reasoning process according to the requirements. \textbf{Your output should be as detailed as possible.} Don't output any other text.
Requirements: Please reason step by step, and put your final answer within \verb|\boxed{}|.\\
The question: <QUESTION>\\
The correct answer: <ANSWER>
\end{minipage}
\\
\midrule

Standard CoT &
\begin{minipage}[t]{\linewidth}\vspace{0pt}\ttfamily\footnotesize
You are a mathematics expert. Now I will present you with a mathematical problem and its correct answer. You need to output the correct reasoning process according to the requirements. Don't output any other text.
Requirements: Please reason step by step, and put your final answer within \verb|\boxed{}|.\\
The question: <QUESTION>\\
The correct answer: <ANSWER>
\end{minipage}
\\
\midrule

Concise CoT &
\begin{minipage}[t]{\linewidth}\vspace{0pt}\ttfamily\footnotesize
You are a mathematics expert. Now I will present you with a mathematical problem and its correct answer. You need to output the correct reasoning process according to the requirements. \textbf{Your output should be as brief and concise as possible.} Don't output any other text.
Requirements: Please reason step by step, and put your final answer within \verb|\boxed{}|.\\
The question: <QUESTION>\\
The correct answer: <ANSWER>
\end{minipage}
\\
\midrule

Ultra-Concise CoT &
\begin{minipage}[t]{\linewidth}\vspace{0pt}\ttfamily\footnotesize
You are a mathematics expert. Now I will present you with a reasoning text related to a mathematical problem. \textbf{You need to rephrase the text according to the requirements. Your output should be as brief and concise as possible.} Don't output any other text.\\
Requirements: Please reason step by step, and put your final answer within \verb|\boxed{}|.\\
The question: <QUESTION>\\
The text: <REFERENCE COT>
\end{minipage}
\\

\bottomrule
\end{tabularx}

\caption{Four prompt templates used in our pipeline. Placeholders: \texttt{<QUESTION>} for the input question, \texttt{<ANSWER>} for the reference answer, and \texttt{<REFERENCE COT>} for the human-written concise reference trace.}
\label{tab:prompt_templates}
\end{table*}

\section{CoT Recovery}

\label{app:cot_recovery}

\begin{figure}[H]       % [t]=顶部 [h]=当前位置 [b]=底部 [!ht]=尽量当前位置
  \centering
  \includegraphics[width=\linewidth]{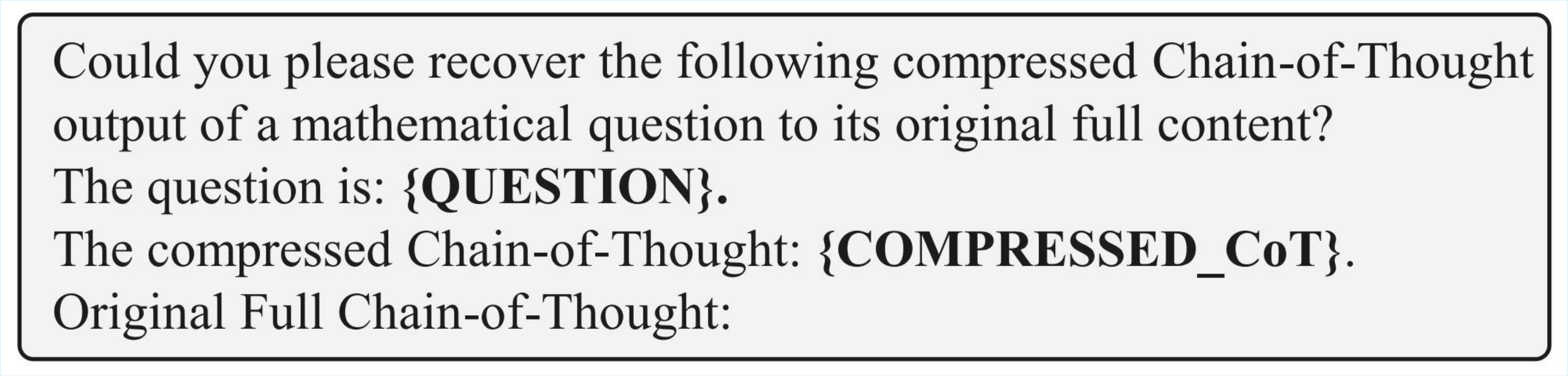} % 也可以写成 width=0.8\linewidth
  \caption{Recovery prompt template for expanding compressed CoTs.}
  \label{fig:recover_prompt}
\end{figure}

Our method performs token-level compression of reasoning traces, which may reduce their readability. 
We follow \citep{xia2025tokenskip} and use a dedicated recovery prompt to expand a compressed CoT into a complete CoT. 
Figure~\ref{fig:recover_prompt} shows the recovery template used in our experiments. 
Given a compressed CoT $\tilde{r}$, we prompt the model to rewrite it into a readable full CoT $r^{\text{rec}}$ while preserving the final answer. 
% This demonstrates that, despite token-level pruning, the essential reasoning is preserved and can be recovered by a LLM.

% We follow \citep{xia2025tokenskip} and use a dedicated recovery prompt to expand a compressed rationale into a complete CoT.
% Figure~\ref{fig:recover_prompt} shows the recovery template used in our experiments.
% Given a compressed rationale $\tilde{r}$, we prompt the model to rewrite it into a readable full rationale $r^{\text{rec}}$ while preserving the final answer.

\begin{figure}[H]       % [t]=顶部 [h]=当前位置 [b]=底部 [!ht]=尽量当前位置
  \centering
  \includegraphics[width=\linewidth]{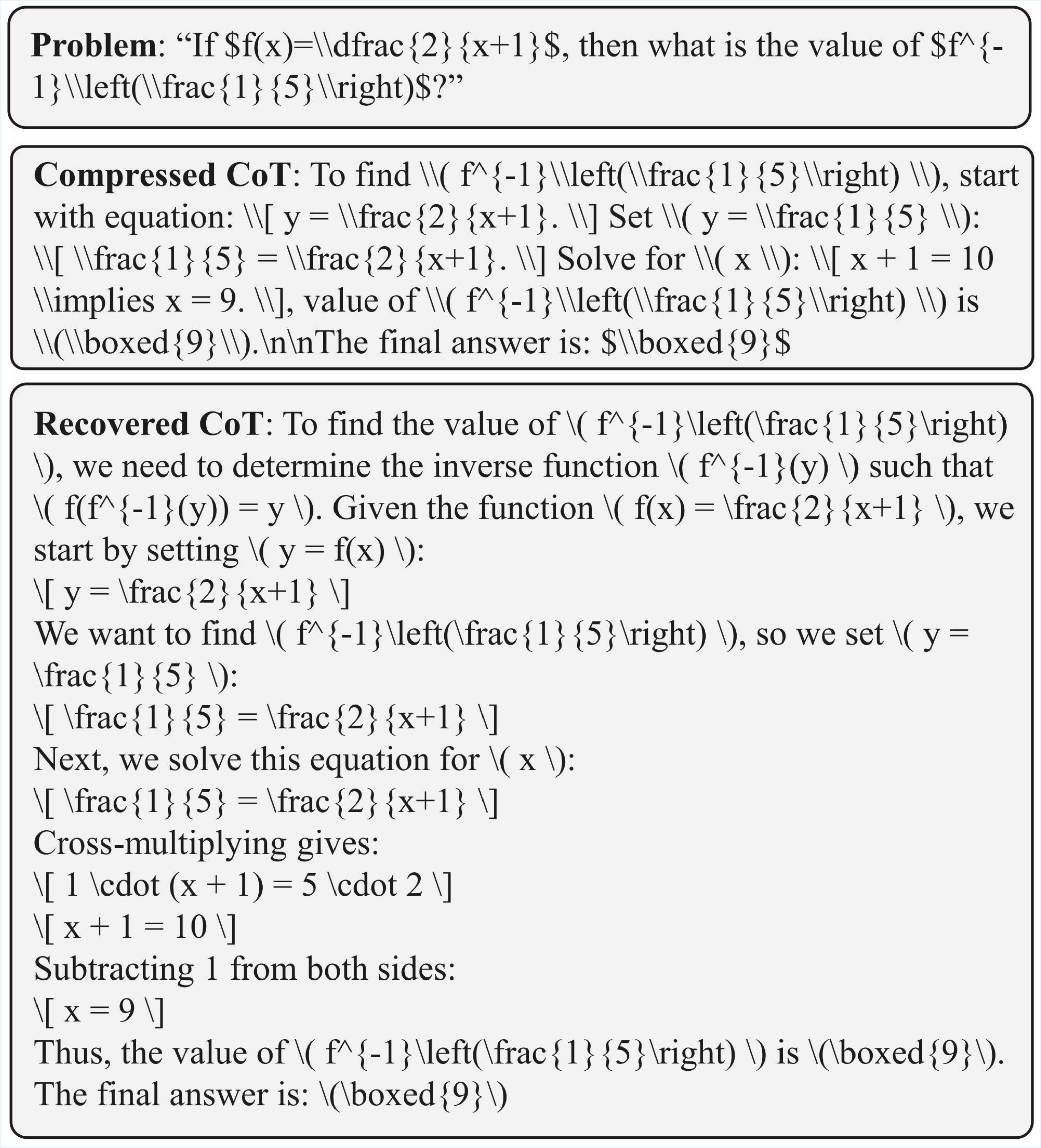} % 也可以写成 width=0.8\linewidth
  \caption{An example of recovering a complete CoT from a compressed CoT.}
  \label{fig:cot_recovery}
\end{figure}

Figure~\ref{fig:cot_recovery} presents a representative recovery example.
The model restores omitted intermediate steps and connective text from the compressed input, and the recovered CoT remains answer-consistent, indicating that the compressed CoT retains sufficient semantic anchors for reconstruction.

\paragraph{Prompt Template.}
To construct the four tiers of Hierarchical Reasoning Abstraction, we use a set of tier-specific prompt templates that control verbosity and step granularity (\textit{Detailed}, \textit{Standard}, \textit{Concise}, \textit{Ultra-Concise}).
For the first three tiers, the prompt provides the input question and the ground-truth answer, and asks the model to produce a supporting CoT with the desired level of detail while keeping the final answer unchanged.
For the \textit{Ultra-Concise} tier, we additionally supply a human-written concise reference trace and instruct the model to produce a minimal CoT that preserves the core logic.
The full prompt templates are provided in Table~\ref{tab:prompt_templates}.

\end{document}